\documentclass[11pt]{article}

\usepackage[preprint]{acl}

\usepackage[T1]{fontenc}

\usepackage[utf8]{inputenc}

\usepackage{inconsolata}

\usepackage{times}
\usepackage{latexsym}
\usepackage{amsmath,amssymb,mathtools}
\usepackage{booktabs}
\usepackage{graphicx}
\usepackage{xcolor}
\usepackage{xspace}
\usepackage{microtype}
\usepackage{url}
\usepackage{hyperref}
\hypersetup{hidelinks}
\usepackage{enumitem}
\usepackage{multirow}
\usepackage{adjustbox}
\newcommand{\method}{\textsc{SRAT}\xspace}
\newcommand{\concat}{\mathbin{\Vert}}

\usepackage{CJKutf8}
\usepackage[most]{tcolorbox}
\usepackage{xcolor}

\newtcolorbox{casebox}[1]{
  enhanced,
  breakable,
  colback=white,
  colframe=black!60,
  colbacktitle=black!10,
  coltitle=black,
  fonttitle=\fontsize{8.5pt}{9.5pt}\selectfont\bfseries,
  fontupper=\fontsize{8.2pt}{9.4pt}\selectfont,
  title=#1,
  boxrule=0.55pt,
  arc=1mm,
  outer arc=1mm,
  left=1.8mm,
  right=1.8mm,
  top=1.3mm,
  bottom=1.3mm,
  before skip=5pt,
  after skip=7pt,
  titlerule=0.45pt
}

\newcommand{\casefield}[2]{%
  \par\noindent\textbf{#1:} #2\par
}

\newcommand{\casebar}[1]{%
  \par\vspace{1.3mm}
  \noindent
  \colorbox{black!7}{%
    \parbox{\dimexpr\linewidth-2\fboxsep\relax}{%
      {\fontsize{8pt}{9pt}\selectfont\bfseries #1}%
    }%
  }%
  \par\vspace{0.8mm}
}

\newcommand{\casescore}[1]{%
  \par\vspace{0.4mm}
  {\fontsize{7.3pt}{8.3pt}\selectfont\itshape #1}
  \par
}

\newcommand{\trad}[1]{{\CJKfamily{bsmi}#1}}

\newtcolorbox{promptbox}[1]{
  enhanced,
  colback=white,
  colframe=black!55,
  colbacktitle=black!8,
  coltitle=black,
  title=#1,
  fonttitle=\fontsize{8.3pt}{9.3pt}\selectfont\bfseries,
  fontupper=\fontsize{7.8pt}{9.0pt}\selectfont\ttfamily,
  width=\linewidth,
  boxrule=0.5pt,
  titlerule=0.4pt,
  arc=0.8mm,
  outer arc=0.8mm,
  left=1.8mm,
  right=1.8mm,
  top=1.2mm,
  bottom=1.2mm,
  before skip=5pt,
  after skip=7pt,
  before upper={
    \raggedright
    \setlength{\parindent}{0pt}
    \setlength{\parskip}{2pt}
  }
}

\newtcolorbox{promptboxcont}[1]{
  enhanced,
  colback=white,
  colframe=black!55,
  colbacktitle=black!6,
  coltitle=black,
  title=#1,
  fonttitle=\fontsize{8.3pt}{9.3pt}\selectfont\bfseries,
  fontupper=\fontsize{7.8pt}{9.0pt}\selectfont\ttfamily,
  width=\linewidth,
  boxrule=0.5pt,
  titlerule=0.4pt,
  arc=0.8mm,
  outer arc=0.8mm,
  left=1.8mm,
  right=1.8mm,
  top=1.2mm,
  bottom=1.2mm,
  before skip=3pt,
  after skip=7pt,
  before upper={
    \raggedright
    \setlength{\parindent}{0pt}
    \setlength{\parskip}{2pt}
  }
}

\definecolor{tagblue}{RGB}{0,0,220}
\definecolor{tagcyan}{RGB}{0,150,200}
\definecolor{tagbrown}{RGB}{160,100,40}
\definecolor{tagpink}{RGB}{200,0,100}

\newcommand{\thinkopen}{\textcolor{tagblue}{\texttt{\textless think\textgreater}}}

\newcommand{\toolopen}{\textcolor{tagcyan}{\texttt{\textless tool\_call\textgreater}}}

\newcommand{\answeropen}{\textcolor{tagpink}{\texttt{\textless answer\textgreater}}}

\title{Retrieve What Helps, Only When It Helps:
Agentic Reinforcement Learning for Selective
Retrieval-Augmented Machine Translation}

\author{
  Shijun Li$^{1}$\thanks{Work was done while Shijun was interning at Amazon.},
  Yang Liu$^{2}$,
  Najmeh Sadoughi$^{2}$, 
  Dongmei Wang$^{2}$,\\
  \textbf{Xinjie Li}$^{2}$,
  \textbf{Zhu Liu}$^{2}$,
  \textbf{Rohith Mysore Vijaya Kumar}$^{2}$,
  \textbf{Yuzong Liu}$^{2}$
  \\[0.5ex]
  $^{1}$The University of Texas at Austin, $^{2}$Amazon\\
  \texttt{shijunli@utexas.edu}\\
}
\begin{document}
\maketitle

\begin{abstract}

Retrieving external knowledge can improve large language model (LLM) translation by providing supplementary information such as idiom interpretations and terminology. However, existing retrieval-augmented machine translation systems typically apply predefined retrieval strategies uniformly across all inputs. We argue that translation operates in a {partial-evidence regime}: LLM's parametric knowledge suffices for many inputs, while external evidence benefits a subset and usually supports rather than directly provides the desired translation. This creates two challenges: deciding \emph{when} to retrieve and \emph{what} evidence to acquire. We introduce Selective Retrieval-Augmented Translation (\method), an agentic reinforcement learning framework that learns when to translate directly or retrieve, what to retrieve, and how to integrate these decisions by optimizing final translation quality. To align evidence retrieval with downstream translation quality rather than simple semantic similarity, we further develop joint Plackett--Luce optimization and translation-quality contrastive training for retriever-training. Experiments on idiom, knowledge-intensive, and document-level translation with offline or online retrieval show that \method consistently outperforms never- and always-retrieve baselines, and quality-aligned retriever training provides further gains when similarity-based retrieval is insufficient. The results demonstrate the benefit of treating retrieval as a learnable translation action rather than fixed preprocessing. 

\end{abstract}

\section{Introduction}
\label{sec:intro}

Large language models (LLMs) have become strong general-purpose machine
translation (MT) systems, particularly after multilingual instruction tuning
and translation-specific adaptation
\citep{jiao-etal-2023-chatgpt,xuetal2024alma}.
In particular, recent LLMs demonstrate strong ability to interpret contextual cues and leverage external knowledge. These capabilities have motivated retrieval-augmented MT systems that provide bilingual examples, idiom definitions, terminology, or supporting documents to further support and improve translation quality
\citep{li-etal-2024-idiomkb,wang-etal-2025-retrieval}.

\begin{figure}[t]
    \centering
    \includegraphics[width=\linewidth]{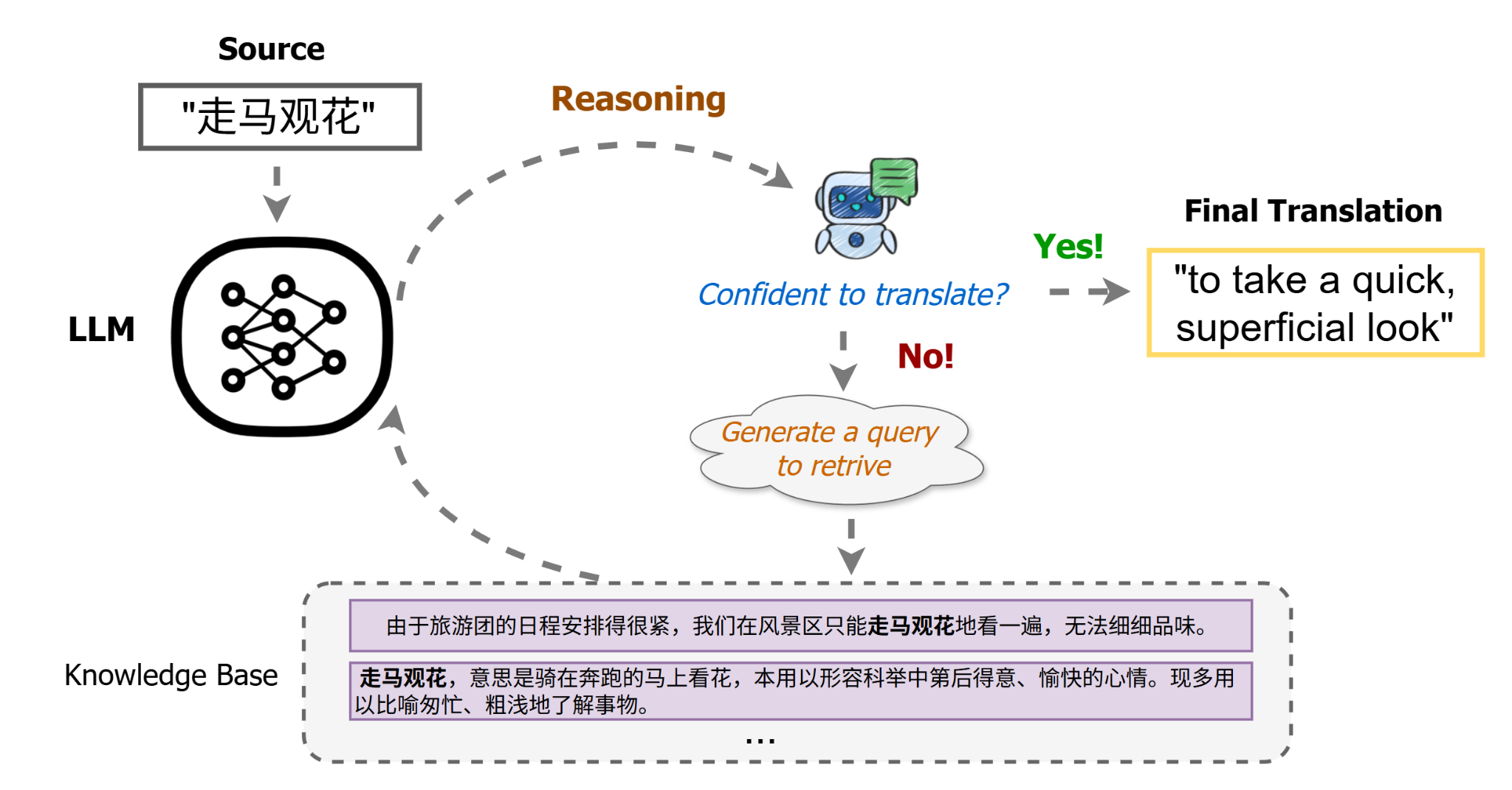}
    \caption{SRAT's workflow by integrating reasoning, retrieval, and translation for LLM-based translation.}
    \label{fig:agentic}
\end{figure}

However, recent studies \citep{sun2026exposing, wang2023document} have shown that retrieval is not uniformly beneficial for translation. Existing
retrieval-augmented MT pipelines typically apply a predefined retrieval strategy: retrieve for every input  with a heuristic, similarity-based matching rule. This overlooks a fundamental gap in LLM-based translation: for some inputs, the model's parametric knowledge is already sufficient, while for others, external evidence such as idiom interpretations, canonical entities, technical terms, or background knowledge can be crucial. Moreover, heuristic, similarity-based retrieval can provide no help and even introduce noisy or misleading context that distracts the model and degrades an otherwise correct translation \citep{sun2026exposing}. For example, a semantically close document may merely repeat query terms without providing actionable knowledge, while a more distant one may contain exactly the idiom interpretation, entity convention, or background knowledge the translation requires. Translation therefore differs from answer-oriented retrieval settings, in which a relevant document often directly contains the evidence needed to produce the final response.


We refer to this setting as the {partial-evidence regime}, which captures two properties of retrieval in MT. First, external evidence is beneficial for only a subset of inputs where the model's parametric knowledge doesn't suffices. Second, when evidence is needed, retrieval based solely on semantic similarity may fail to surface the most useful evidence.
This raises two learning problems. The first is \emph{retrieval adaptivity}: the model must determine \emph{when} additional evidence is useful instead of retrieving by default. The second is \emph{retrieval quality alignment}: once retrieval is invoked, the system must determine {what} evidence actually improves the final translation quality.

To address them, we propose Selective Retrieval-Augmented Translation (\method), an agentic reinforcement-learning framework that treats reasoning, retrieval, and translation as actions in a unified policy. Given a source input, the model's action space at each step includes reasoning about the translation, generating a retrieval query and incorporating the returned evidence, or terminating with the final translation.  Rather than supervising which examples should retrieve, we optimize complete trajectories using final translation quality. Group Relative Policy Optimization (GRPO) \citep{shaoetal2024deepseekmath} allows trajectories with different retrieval decisions and queries to compete for the same input, thereby learning when retrieval is worthwhile and what to retrieve to improve final translation quality.

Moreover, we develop two approaches for further aligning the retriever with downstream translation quality rather than simple semantic similarity. First, a natural and principled approach is to jointly optimize the LLM translator and retriever end-to-end. To achieve it, we replace deterministic top-$k$ selection during training with a Plackett--Luce ranking policy \citep{plackett1975rankings}, whose selection log-probability allows trajectory-level translation reward to update the retriever through a score-function gradient. However, this joint signal can be noisy: a trajectory contains many LLM policy decisions but only a few retrieval actions, making credit assignment to the retriever inherently high-variance. To address this, we further introduce translation-quality contrastive training, which fixes the translator, evaluates candidate retrieval evidence by their measured effect on translation quality, and converts these into direct contrastive supervision. This isolates the effect of retrieval from other policy decisions, providing the retriever with substantially more targeted training signals.

We evaluate \method across three translation settings: idiom
translation, knowledge-intensive translation, and document-level translation. The first two use offline retrieval corpora, while the document-level setting additionally evaluates \method with online retrieval. Across these settings, \method consistently outperforms
never-retrieve and always-retrieve baselines, demonstrating the benefit of learning when to retrieve and what evidence to seek. Ablations  show that quality-aligned retriever training provides further significant gains when similarity-based retrieval is insufficient. We also find that the retrieval policy can be optimized with a frontier black-box translator, outperforming its direct translation and indicating that the adaptive retrieval decision training can be decoupled from the translator and optimized as a modular component for external generation models.

Our main contributions are threefold:

\begin{itemize}[leftmargin=*]

    \item \textbf{Selective retrieval as a translation decision.}
    We characterize LLM-based RAG translation as a partial-evidence regime in
    which external knowledge is useful only conditionally. We formulate machine translation by LLM as a sequential policy over reasoning, retrieval, and final translation, enabling
    the model to learn \emph{when} and \emph{what} to retrieve from
    translation outcomes rather than fixed routing rules.

    \item \textbf{Quality-aligned evidence retrieval.}
    We propose two retriever-training strategies. Joint Plackett–Luce policy-gradient training propagates trajectory-level translation reward to evidence retrieval, while translation-quality contrastive training derives direct retriever supervision from controlled retrieval evidence interventions. Both align the retriever toward the downstream effect of retrieved evidence on translation quality rather than semantic similarity alone.

    \item \textbf{Evaluation across retrieval regimes.}
    Experiments on idiom, knowledge-intensive, and document-level translation with offline and online retrieval consistently outperform fixed never- and always-retrieve strategies. The results further show that quality-aligned retrieval is significantly beneficial when similarity-based retrieval is insufficient, and that adaptive retrieval policy learning can be decoupled from the translator to improve frontier black-box models beyond their direct translation performance.
    
\end{itemize}

\section{Related Work}
\label{sec:related}

\noindent \textbf{Large Language Models for Translation.}
Early studies showed that general-purpose LLMs can deliver competitive
translation performance, although results vary across languages, domains, and
prompting strategies
\citep{jiao-etal-2023-chatgpt,hendyetal2023gptmt}. Translation-specific
continued pretraining and supervised fine-tuning further improve smaller open
models; ALMA, for example, combines monolingual adaptation with high-quality
parallel data \citep{xuetal2024alma}. These efforts improve the underlying
translator, but largely leave inference-time information acquisition outside
the modeling problem. A parallel line of work introduces explicit deliberation into LLM-based MT.
MAPS elicits multiple translation-relevant analyses before selecting among
candidate outputs \citep{he-etal-2024-maps}; TEaR adopts a
translate--estimate--refine procedure \citep{feng-etal-2025-tear}; and DRT
distills extended reasoning traces for figurative text
\citep{wang-etal-2025-drt}. These studies suggest that additional computation
can help difficult inputs, but the form and amount are typically fixed in
advance. \method instead learns when extra reasoning and retrieval really help for translation.

\noindent \textbf{Retrieval-Augmented Machine Translation.}
Retrieval has long been used in MT through translation memories and
nearest-neighbor examples, while recent LLM-based systems draw on broader forms
of external evidence. IdiomKB retrieves explanatory knowledge for idiomatic
translation \citep{li-etal-2024-idiomkb}, RAGtrans incorporates multilingual
unstructured documents \citep{wang-etal-2025-retrieval}, and cross-cultural MT
has explored multilingual knowledge graphs
\citep{conia-etal-2024-cross-cultural}. These results demonstrate the value of
external knowledge when the source alone is insufficient. Yet most systems
either retrieve by default or rely on predefined routing, while retrieval is
often optimized separately from the final translation objective. Our focus is
instead on whether retrieved evidence improves translation quality for a given
input and model state.

\noindent \textbf{Adaptive Retrieval and Reinforcement Learning.}
Adaptive retrieval has been studied mainly in language modeling and question
answering. FLARE triggers retrieval when generation becomes uncertain
\citep{jiang-etal-2023-flare}; Self-RAG learns retrieval and reflection actions
from supervised preference data \citep{asaietal2024selfrag}; and Search-R1 uses
reinforcement learning to acquire multi-turn search-and-reason behavior from
answer-level rewards \citep{jinetal2025searchr1}. These works show that
retrieval can be integrated into a learned generation policy. Translation,
however, differs in that a plausible output can often be produced without
retrieval, while useful evidence may guide the rendering without explicitly
containing the target translation. This makes selective retrieval and alignment
with translation quality especially important.

\section{Method}
\label{sec:method}

Our method addresses two decisions induced by the partial-evidence regime:
\emph{when} external evidence should be acquired and \emph{what} evidence
should be returned when retrieval is invoked. We first formulate selective
retrieval as part of the translation policy, and then introduce two strategies
for aligning the retriever with downstream translation quality.

\subsection{Selective Retrieval as a Translation Policy}
\label{sec:policy-formulation}

Let $x$ denote a source sentence, $P(\ell_s,\ell_t)$ the translation instruction prompt given source and target languages $\ell_s$ and $\ell_t$, and
$y^\star$ a reference available only during training. An external corpus
$\mathcal{D}=\{d_j\}_{j=1}^{N}$ contains potentially useful evidence such as
idiom explanations, terminology and entity,  background documents, or online information.

We model the LLM translator as a policy $\pi_\theta$ with initial state
$s_0=[P(\ell_s,\ell_t),x]$. Starting from $s_0$, the policy generates a
trajectory $\tau=(a_1,\ldots,a_T)$ that may interleave reasoning, retrieval,
and the final translation $\hat y$. Making information acquisition part of the
trajectory allows retrieval to be decided dynamically rather than applied as a fixed preprocessing step.
Specifically, the trajectory contains three types of action during the workflow:
\begin{itemize}[leftmargin=*]
    \item \textbf{Reasoning (\thinkopen{}).}
    The model may analyze ambiguity, figurative meaning, cultural context, entities, or terminology, or assess whether its current knowledge is sufficient for translation.

    \item \textbf{Retrieval (\toolopen{}).}
    When the current evidence
    is insufficient, the model generates a natural-language query $q_t$ and  invokes the retriever $r_\phi$ over $\mathcal{D}$:
    \begin{equation}
        D_t\sim r_\phi(\cdot\mid q_t,\mathcal{D}).
    \end{equation}
    The returned document set is appended to the interaction state,
    $s_{t+1}=s_t\concat[q_t,D_t]$, after which the model can reason the evidence, regenerate a query, or directly translate. The query can directly target the missing information, for example, the meaning of an ambiguous expression in context or relevant background knowledge needed for translation.

    \item \textbf{Translation (\answeropen{}).}
    Once the model judges the available evidence sufficient, it emits final translation
    $\hat y$ and
    terminates.  This action can occur immediately after the initial state, so direct translation is a valid zero-retrieval trajectory.  
\end{itemize}

The policy therefore jointly learns \emph{whether} to retrieve, \emph{what}
information to request, and \emph{how} to use all evidence for translation. The
objective is not to apply a fixed retrieval policy, but to retrieve only when it
improves the final translation.

\subsection{Learning When to Retrieve}
\label{sec:policy-learning}

\textbf{Outcome-guided Reinforcement Learning.}
We train the policy from a trajectory-level reward:
\begin{equation}
\label{eq:reward}
R(\tau)
=
Q(\hat y,y^\star,x)
-\lambda_{\mathrm{fmt}}M(\tau)
-\lambda_{\mathrm{cost}}C(\tau),
\end{equation}
where $Q$ measures translation quality, $M(\tau)$ penalizes malformed outputs,
and $C(\tau)$ penalizes excessive redundancy or repetition in the generation. In our experiments,
$Q$ combines COMET \citep{rei-etal-2020-comet} with an LLM-based translation evaluator.

For each source, we sample a group of $G$ trajectories and compute the group advantage:
\begin{equation}
\label{eq:advantage}
A_i=
\frac{R(\tau_i)-\overline R}
{\operatorname{std}(R(\tau_1),\ldots,R(\tau_G))+\epsilon}.
\end{equation}
We then optimize $\pi_\theta$ using the standard clipped GRPO objective
\citep{shaoetal2024deepseekmath}. Because trajectories for the same input can
differ in whether they retrieve, what they query, and how they make final translation,
their relative rewards directly supervise selective retrieval without
requiring labels specifying which inputs need external knowledge.

\textbf{Retrieval exploration curriculum.}
Optimizing the outcome reward directly from an off-the-shelf model can collapse to a never-retrieve policy: early in training, the model's self-generated queries are often poor and its retrieval-action formatting is frequently malformed, so direct translation consistently receives higher reward before useful retrieval behavior has a chance to emerge.
We therefore use a three-stage curriculum. First, we explicitly generate
retrieval-augmented trajectories and retain the top 40\% according to quality $Q$. Second, we perform supervised fine-tuning on these trajectories to initialize the interaction format and query-generation ability. Third, we conduct RL with
a temporary retrieval penalty to move the SFT policy away from always
retrieving, and then remove this shaping term and optimize
Eq.(\ref{eq:reward}) directly until convergence. The final policy is therefore free to retrieve or translate directly according to downstream quality rather than a predefined retrieval rate.

\subsection{Learning What to Retrieve}
\label{sec:quality-retriever}

The policy above learns whether to retrieve and how to formulate queries, but a fixed retriever still ranks documents according to its pretraining objective, typically the shallow semantic similarity. Since our objective is to retrieve documents to help make better translation, we instead optimize document selection according
to translation quality. We consider two complementary approaches.

\subsubsection{Joint Policy-Gradient Alignment}
\label{sec:joint}

Given query $q$ and candidate set $\mathcal{C}_q$, the retriever assigns each
document a score $z_j=f_\phi(q,d_j)$. Deterministic top-$k$ selection does not provide a selection log-probability through which the translation reward can
update the retriever. We instead define a stochastic ranking policy.  Given a query $q$ and candidate
set $\mathcal{C}_q$, the retriever assigns logits
$z_j=f_\phi(q,d_j)$.  An ordered sample
$S=(d^{(1)},\ldots,d^{(k)})$ drawn without replacement follows the Plackett--Luce distribution with temperature $T$ as:
\begin{equation}
\label{eq:pl}
p_\phi(S\mid q)=\prod_{m=1}^{k}
\frac{\exp(z_{d^{(m)}}/T)}
{\sum_{d\in\mathcal{C}_q\setminus S_{<m}}\exp(z_d/T)},
\end{equation}
where $\mathcal{C}_q\setminus S_{<m}$ denotes the remaining candidates after the first $m-1$ selections. This stochastic sampling makes the selection a differentiable distribution even though each draw is discrete.

For a trajectory containing retrieval events $\mathcal{R}(\tau)$, the joint
log probability factorizes as:
\begin{equation}
\begin{aligned}
\log p_{\theta,\phi}(\tau)=
\sum_t & \log\pi_\theta(a_t\mid s_t) + \\
&\sum_{r\in\mathcal{R}(\tau)}\log p_\phi(S_r\mid q_r).
\end{aligned}
\end{equation}

Using the group advantage from Eq.(\ref{eq:advantage}), the retriever
gradient is:
\begin{equation}
\label{eq:retgrad}
\nabla_\phi J
=\mathbb{E}_{\tau}\!\left[
A(\tau)\sum_{r\in\mathcal{R}(\tau)}
\nabla_\phi\log p_\phi(S_r\mid q_r)
\right].
\end{equation}
The same outcome therefore increases the probability of documents selected in high-advantage trajectories and decreases it in low-advantage trajectories.
Importantly, this optimizes quality under the current translator and policy; it does not assume that semantic similarity is itself the target.

\begin{figure}[t]
    \centering
    \includegraphics[width=0.85\linewidth]{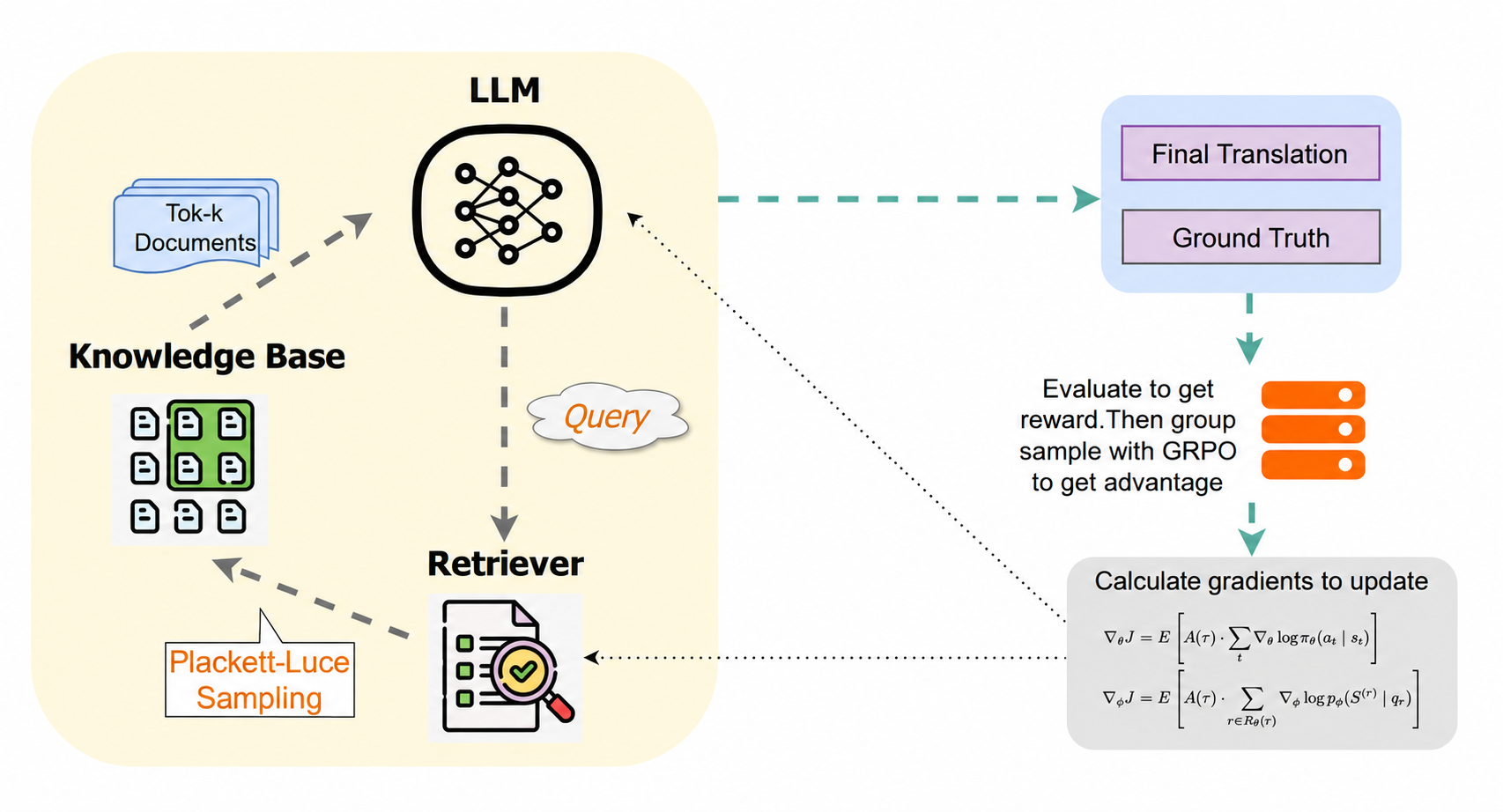}
    \caption{
    Joint translator--retriever optimization. Plackett--Luce sampling exposes
    the retriever selection probability, allowing the trajectory advantage to
    update both components.
    }
    \label{fig:joint}
\end{figure}

\subsubsection{Separate Contrastive Alignment}
\label{sec:contrastive}

Joint optimization using Eq.(\ref{eq:retgrad}) may suffer from high variance in credit assignment to the retriever: a single terminal reward must be attributed across many token-level decisions but only a handful of document selections, diluting the gradient signal that reaches the retriever parameters. We therefore introduce a
complementary contrastive approach that provides direct document-level
supervision.

We first train $\pi_\theta$ with a fixed retriever. For each query $q$ produced by the trained policy on the training data, we retrieve a candidate set $\mathcal{C}_q=\{d_1,\ldots,d_M\}$ and run controlled translations using one candidate at a time. We score the resulting translations with the same quality function $Q$ and choose the document producing the best translation as
$d^+=\arg\max_{d_i\in\mathcal{C}_q}Q(\hat y_i,y^\star,x)$.

We then fine-tune the retriever using:
\begin{equation}
\label{eq:infonce}
\mathcal{L}_{\mathrm{ret}}(\phi)
=
-\log
\frac{\exp(f_\phi(q,d^+)/T_c)}
{\sum_{d_i\in\mathcal{C}_q}\exp(f_\phi(q,d_i)/T_c)}.
\end{equation}
Unlike conventional contrastive retrieval, the positive document is defined by
its measured effect on translation rather than annotated relevance or embedding
similarity. This gives the retriever a focused, lower-variance signal about
which evidence the current translator can actually use.

\begin{figure}[t]
    \centering
    \includegraphics[width=\linewidth]{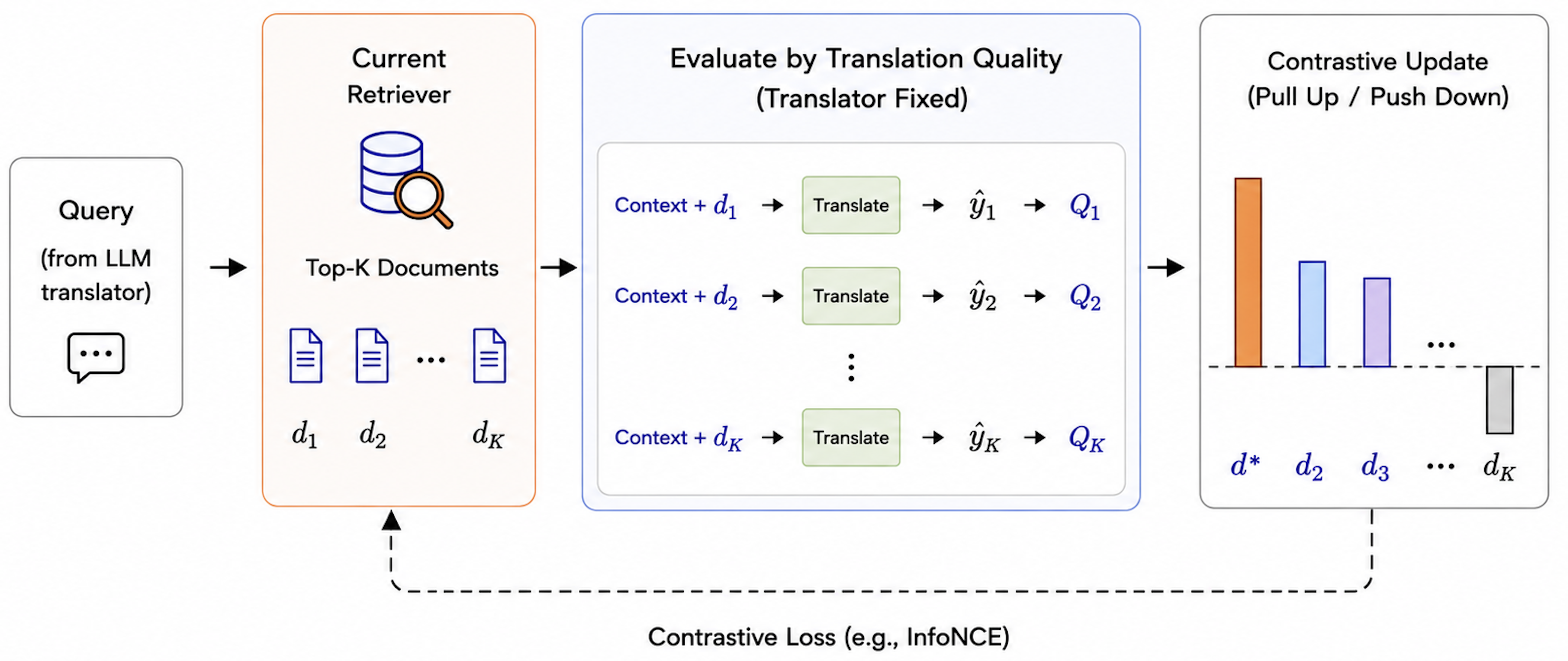}
    \caption{
    Translation-quality contrastive alignment. Candidate documents are evaluated
    by their effect on translation quality and converted into direct retriever
    supervision with the translator fixed.
    }
    \label{fig:contrastive}
\end{figure}

These two strategies offer complementary trade-offs. Joint optimization updates
the translator and retriever in an end-to-end way, while contrastive
alignment requires additional candidate evaluations but provides more direct
retriever supervision. We compare and discuss  them in
Section~\ref{sec:rq2}.

\subsection{Inference}
\label{sec:inference}

At inference time, no reference translation or quality evaluator is required. The learned policy decides whether retrieval is necessary, generates a query when needed and incorporates the returned evidence, and produces the final translation. 
We set a maximum number of retrieval calls and an output-token budget to bound worst-case latency.  We report both translation quality and retrieval rate to measure the effectiveness and policy behaviors.

\section{Experiments}
\label{sec:experiments}

\begin{table*}[t]
\centering
\scriptsize
\setlength{\tabcolsep}{2.0pt}
\renewcommand{\arraystretch}{0.92}
\resizebox{0.92\linewidth}{!}{%
\begin{tabular}{lrrrrrrrrrrrr}
\toprule
& \multicolumn{6}{c}{Zh$\rightarrow$En}
& \multicolumn{6}{c}{En$\rightarrow$Zh} \\
\cmidrule(lr){2-7}
\cmidrule(lr){8-13}
Model
& BLEU & COMET & I5 & I3 & MQM & Ret.\,(\%)
& BLEU & COMET & I5 & I3 & MQM & Ret.\,(\%) \\
\midrule
Direct
& 21.50 & 0.805 & 0.652 & 0.668 & -3.94 & 0.0
& 33.10 & 0.829 & 0.646 & 0.689 & -3.88 & 0.0 \\
Self-CoT
& 20.18 & 0.795 & 0.673 & 0.658 & -5.10 & 0.0
& 33.02 & 0.838 & 0.650 & 0.708 & -3.32 & 0.0 \\
KB-CoT
& 19.44 & 0.792 & 0.658 & 0.644 & -4.50 & 100.0
& 33.86 & 0.842 & 0.662 & 0.700 & -3.84 & 100.0 \\
Ladder
& 24.36 & 0.804 & 0.646 & 0.637 & -3.60 & 0.0
& 47.33 & 0.857 & 0.750 & 0.814 & -2.30 & 0.0 \\
CSC
& \textbf{24.95} & 0.812 & 0.635 & 0.643 & -4.01 & 100.0
& \textbf{48.47} & 0.860 & 0.763 & 0.829 & -2.11 & 100.0 \\
\midrule
\textsc{SRAT-N}
& 24.41 & 0.810 & 0.709 & 0.693 & -3.02 & 0.1
& 44.44 & 0.878 & 0.746 & 0.807 & -2.25 & 0.0 \\
\textsc{SRAT-A}
& 22.48 & 0.808 & 0.700 & 0.686 & -3.45 & 100.0
& 44.49 & 0.880 & 0.754 & 0.825 & -2.24 & 100.0 \\
\textsc{SRAT}
& 24.62 & \textbf{0.822} & \textbf{0.731} & \textbf{0.728}
& \textbf{-2.74} & 30.6
& 48.25 & \textbf{0.891} & \textbf{0.792} & \textbf{0.871}
& \textbf{-1.85} & 77.0 \\
\bottomrule
\end{tabular}}
\caption{
Results on IdiomKB. I5 and I3 denote Idiom-Judge-5 and
Idiom-Judge-3.
}
\label{tab:idiom-main}
\end{table*}

\begin{table*}[t]
\centering
\scriptsize
\setlength{\tabcolsep}{2.0pt}
\renewcommand{\arraystretch}{0.92}
\resizebox{0.92\linewidth}{!}{%
\begin{tabular}{lrrrrrrrrrrrr}
\toprule
& \multicolumn{6}{c}{En$\rightarrow$Zh}
& \multicolumn{6}{c}{En$\rightarrow$De} \\
\cmidrule(lr){2-7}
\cmidrule(lr){8-13}
Model
& BLEU & COMET & GRF & GRB & MQM & Ret.\,(\%)
& BLEU & COMET & GRF & GRB & MQM & Ret.\,(\%) \\
\midrule
Direct
& 45.74 & 0.867 & 0.856 & 0.774 & -11.27 & 0.0
& 29.40 & 0.806 & 0.743 & 0.654 & -21.28 & 0.0 \\
Self-CoT
& 49.67 & 0.872 & 0.870 & 0.781 & -10.07 & 0.0
& 29.95 & 0.819 & 0.775 & 0.681 & -19.27 & 0.0 \\
KB-CoT
& 46.01 & 0.868 & 0.874 & 0.786 & -11.60 & 100.0
& 29.89 & 0.808 & 0.747 & 0.658 & -20.72 & 100.0 \\
Ladder
& 54.05 & 0.875 & 0.913 & 0.831 & -5.50 & 0.0
& \textbf{38.38} & 0.829 & 0.877 & 0.779 & -7.93 & 0.0 \\
CSC
& \textbf{55.42} & 0.880 & 0.920 & 0.840 & -5.40 & 100.0
& 38.18 & 0.836 & 0.880 & 0.796 & -7.54 & 100.0 \\
\midrule
\textsc{SRAT-N}
& 49.32 & 0.870 & 0.918 & 0.827 & -5.10 & 0.0
& 36.21 & 0.831 & 0.878 & 0.801 & -7.14 & 0.0 \\
\textsc{SRAT-A}
& 49.32 & 0.876 & 0.923 & 0.835 & -4.98 & 100.0
& 36.12 & 0.838 & 0.884 & 0.793 & -6.86 & 100.0 \\
\textsc{SRAT}
& 52.37 & \textbf{0.888} & \textbf{0.939} & \textbf{0.860}
& \textbf{-4.58} & 52.2
& 36.84 & \textbf{0.857} & \textbf{0.917} & \textbf{0.826}
& \textbf{-5.98} & 43.0 \\
\bottomrule
\end{tabular}}
\caption{
Results on RAGtrans. GRF and GRB are the reference-free and reference-based LLM judges.
}
\label{tab:rag-main}
\end{table*}

We evaluate \method on idiom, knowledge-intensive, and document-level translation with either offline and online retrieval. We conduct experiment to study three research questions:
\textbf{RQ1.} 
Does a learned selective-retrieval policy outperform prior translation methods and matched policies that never or always retrieve?
\textbf{RQ2.} Does aligning the retriever with downstream translation utility improve translation quality, and when is it most beneficial?
\textbf{RQ3.} Can the adaptive retrieval policy be trained with a stronger black-box LLM like Claude as the translator and further improve over its direct translation performance?

\subsection{Experimental Setup}
\label{sec:exp-setup}

\paragraph{Datasets.}
\textbf{IdiomKB} \citep{li-etal-2024-idiomkb} evaluates idiom translation
with short idiom explanations as the retrieval corpus. We use
Zh$\rightarrow$En (8K training / 500 test) and En$\rightarrow$Zh
(4.6K / 500).
\textbf{RAGtrans} \citep{wang-etal-2025-retrieval} evaluates
knowledge-intensive translation with multilingual unstructured documents. We
use En$\rightarrow$Zh (74.5K / 2K) and En$\rightarrow$De
(85.2K / 2K).
\textbf{Document-MT-LLM} \cite{wang2023document} evaluates document-level En$\rightarrow$Ru subtitle
translation (5.5K/500). Unlike the offline-corpus settings, it has no
predefined retrieval collection; the policy instead generates queries for an
online search engine (e.g., DuckDuckGo) to acquire contextual information on demand. These datasets cover structured offline retrieval, open-domain document
retrieval, and online search for long-form translation.

\paragraph{Baselines.}
All open-model systems use Qwen2.5-7B-Instruct as the default backbone.
\textbf{Direct} translates without explicit reasoning or retrieval.
\textbf{Self-CoT} reasons without external evidence, while
\textbf{KB-CoT} \citep{li-etal-2024-idiomkb} conditions reasoning for translation on retrieved knowledge for every case.
\textbf{Ladder} is a retrieval-free baseline fine-tuned for translation
\citep{feng-etal-2024-ladder}.
\textbf{CSC} further fine-tunes the translator with retrieved context under a fixed
always-retrieve pipeline \citep{wang-etal-2025-retrieval}.

We compare the full adaptive policy with two matched policy ablations.
\textbf{\textsc{SRAT-N}} denotes outcome-based RL that converges to a
near-zero retrieval policy, while \textbf{\textsc{SRAT-A}} adds a consistent retrieval
bonus and converges to always retrieving. The full \textbf{\method} model uses
the retrieval-exploration curriculum and is ultimately optimized without a
target retrieval rate. For retriever alignment, \textbf{Joint} denotes
Plackett--Luce joint translator--retriever training, while
\textbf{Contrastive} denotes translation-utility contrastive retriever
training.

\paragraph{Metrics.}
We report BLEU \citep{papineni-etal-2002-bleu}, COMET
\citep{rei-etal-2020-comet}, MQM, and task-specific LLM-judge metrics.
MQM exactly follows the LLM-judge protocol of \citet{zhan2026large}.
For IdiomKB, I3 follows the three-level idiom-translation judge from
\citet{li-etal-2024-idiomkb}, and we extend it to a finer-grained five-level
criterion, noted as I5. For RAGtrans, GRF and GRB exactly follow its reference-free and
reference-based LLM-judge templates \citep{wang-etal-2025-retrieval}.
All LLM-based evaluations use Claude Sonnet 4.6 as default. Higher is better for all metrics; MQM is reported as a negative error score, so values closer to zero
are better. We also report the percentage of test examples where there exists retrieval action.

\paragraph{Training details.}
The RL reward is $R=Q-0.5\times M-0.05\times C$, where
$Q=0.5\times Q_{\mathrm{LLM}}+0.5\times Q_{\mathrm{COMET}}$,
$M$ is a format penalty (1 for malformed outputs and 0 otherwise), and
$C=max(0,(L_{\mathrm{out}}-256)/512)$ measures excessive generation cost.
We use I5 as $Q_{\mathrm{LLM}}$ on IdiomKB and GRF on RAGtrans and Document-MT.
Training uses VeRL with an SGLang backend and a Sentence Transformer \citep{reimers2019sentence} retriever,
with top-$k=1$ as default, at most five interaction turns, and a maximum output length of
2048 tokens. Additional experimental details, qualitative cases, and LLM-Judge prompts are provided in
Appendix \ref{sec:expdetail} and \ref{app:cases}. We also conduct ablations to study the effect of different components like LLM backbones and evaluators in Appendix \ref{app:robustness}.

\begin{table*}[t]
\centering
\scriptsize
\setlength{\tabcolsep}{2.0pt}
\renewcommand{\arraystretch}{0.92}
\resizebox{0.92\linewidth}{!}{%
\begin{tabular}{llrrrrrlrrrrrr}
\toprule
& \multicolumn{6}{c}{IdiomKB}
& \multicolumn{6}{c}{RAGtrans} \\
\cmidrule(lr){2-7}
\cmidrule(lr){8-13}
Alignment
& Dir. & BLEU & COMET & I5 & I3 & MQM
& Dir. & BLEU & COMET & GRF & GRB & MQM \\
\midrule
Similarity
& Zh$\rightarrow$En
& 24.62 & 0.822 & \textbf{0.731} & 0.728 & -2.74
& En$\rightarrow$Zh
& 52.37 & 0.888 & 0.939 & 0.860 & -4.58 \\
Joint
& Zh$\rightarrow$En
& 24.71 & 0.818 & 0.725 & 0.733 & -2.80
& En$\rightarrow$Zh
& 53.49 & 0.891 & 0.941 & 0.863 & -4.41 \\
Contrastive
& Zh$\rightarrow$En
& \textbf{24.75} & \textbf{0.824} & 0.730 & \textbf{0.734}
& \textbf{-2.70}
& En$\rightarrow$Zh
& \textbf{53.99} & \textbf{0.898} & \textbf{0.947}
& \textbf{0.871} & \textbf{-4.27} \\
\midrule
Similarity
& En$\rightarrow$Zh
& 48.25 & 0.891 & 0.792 & 0.871 & -1.85
& En$\rightarrow$De
& 36.84 & 0.857 & 0.917 & 0.826 & -5.98 \\
Joint
& En$\rightarrow$Zh
& 48.73 & 0.886 & 0.790 & \textbf{0.877} & -1.52
& En$\rightarrow$De
& 36.86 & 0.858 & 0.917 & 0.828 & -5.89 \\
Contrastive
& En$\rightarrow$Zh
& \textbf{48.90} & \textbf{0.893} & \textbf{0.795}
& \textbf{0.877} & \textbf{-1.48}
& En$\rightarrow$De
& \textbf{36.93} & \textbf{0.863} & \textbf{0.922}
& \textbf{0.835} & \textbf{-5.54} \\
\bottomrule
\end{tabular}}
\caption{
Retriever alignment on both datasets. Similarity uses the default retriever; Joint uses Plackett--Luce policy-gradient training; and Contrastive uses  separate, contrastive training.
}
\label{tab:retriever-alignment}
\end{table*}

\subsection{RQ1: Does Selective Retrieval Improve Translation?}
\label{sec:rq1}

Tables~\ref{tab:idiom-main}, \ref{tab:rag-main}, and
\ref{tab:document-mt-full} (see in Appendix \ref{app:document-mt}) show a consistent advantage of selective retrieval across
idiom, knowledge-intensive, and document-level translation. \method \footnote{ \scriptsize {Note we train the LLM by fixing the default retriever here (i.e., Sentence Transformer). We discussed the retriever alignment results in Section \ref{sec:rq2}.}} achieves
the strongest COMET, task-specific LLM-judge, and MQM results across all
settings. Compared with the strongest SFT baselines, \method improves I5 by
0.096 and MQM by 1.27 points on IdiomKB Zh$\rightarrow$En, GRF by 0.019 and
MQM by 0.82 points on RAGtrans En$\rightarrow$Zh, and GRF by 0.020 and MQM
by 0.42 points on Document-MT-LLM. The gains are particularly consistent on
metrics that assess semantic adequacy and multidimensional translation quality.

\paragraph{Learning when to retrieve matters.}

The comparison with \textsc{SRAT-N} and \textsc{SRAT-A} isolates the value of the retrieval decision. Across all three benchmarks, \method consistently
outperforms both never- and always-retrieve policies on semantic and
LLM-judged metrics. Relative to \textsc{SRAT-A}, MQM improves by 20.6\% and
17.4\% on the two IdiomKB directions, by 8.0--12.8\% on RAGtrans, and by
5.4\% on Document-MT-LLM. These gains are achieved with substantially
different learned retrieval rates: 30.6--77.0\% on IdiomKB, around half of
the inputs on RAGtrans, and 88.6\% on Document-MT-LLM. This variation suggests
that the policy adapts retrieval behavior to different translation settings
rather than learning a fixed preference for retrieval. Moreover,
\textsc{SRAT-N} already provides substantial gains over the unfine-tuned
model, while \method further improves upon it, indicating complementary
benefits from RL-based translation learning and selective retrieval.

\paragraph{Why is BLEU lower than supervised baselines?}
Ladder or CSC obtains the highest BLEU across the evaluated settings. This is expected because our RL reward does not include BLEU, but instead directly optimizes COMET and task-specific LLM judges, whereas supervised fine-tuning on reference translations can favor surface forms closer to the reference.
BLEU measures n-gram overlap with a limited set of references
\citep{papineni-etal-2002-bleu} and may under-reward valid paraphrases or more natural translations. Prior studies have therefore cautioned against treating BLEU alone as decisive evidence of translation quality
\citep{callison-burch-etal-2006-evaluating,mathur-etal-2020-tangled}, and
human-preferred translations can receive lower BLEU when their wording differs from literal or translationese references \citep{freitag-etal-2020-human}. We therefore treat BLEU as a complementary lexical metric and place main emphasis on COMET and LLM-based evaluation. Additional analyses and qualitative cases on this are provided in Appendix \ref{app:bleu-cases}.

\begin{table*}[t]
\centering
\scriptsize
\setlength{\tabcolsep}{2.0pt}
\renewcommand{\arraystretch}{0.92}
\resizebox{0.92\linewidth}{!}{%
\begin{tabular}{lrrrrrrrrrrrr}
\toprule
{[IdiomKB]}
& \multicolumn{6}{c}{Zh$\rightarrow$En}
& \multicolumn{6}{c}{En$\rightarrow$Zh} \\
\cmidrule(lr){2-7}
\cmidrule(lr){8-13}
Model
& BLEU & COMET & I5 & I3 & MQM & Ret.\,(\%)
& BLEU & COMET & I5 & I3 & MQM & Ret.\,(\%) \\
\midrule
Claude
& 28.30 & 0.826 & 0.758 & 0.752 & \textbf{-0.92} & 0.0
& 54.63 & 0.903 & 0.812 & 0.878 & -0.99 & 0.0 \\
Claude-Ret
& 28.13 & 0.821 & 0.760 & 0.748 & -0.99 & 3.4
& \textbf{57.21} & 0.902 & \textbf{0.814} & \textbf{0.882}
& -1.12 & 11.8 \\
\textsc{SRAT}-Claude
& \textbf{28.39} & \textbf{0.829} & \textbf{0.763} & \textbf{0.755}
& -0.94 & 1.9
& 56.85 & \textbf{0.907} & \textbf{0.814} & 0.879
& \textbf{-0.97} & 10.2 \\
\midrule
{[RAGtrans]}
& \multicolumn{6}{c}{En$\rightarrow$Zh}
& \multicolumn{6}{c}{En$\rightarrow$De} \\
\cmidrule(lr){2-7}
\cmidrule(lr){8-13}
Model
& BLEU & COMET & GRF & GRB & MQM & Ret.\,(\%)
& BLEU & COMET & GRF & GRB & MQM & Ret.\,(\%) \\
\midrule
Claude
& 53.54 & 0.893 & 0.954 & 0.896 & -1.85 & 0.0
& 47.30 & 0.871 & 0.960 & 0.886 & -1.25 & 0.0 \\
Claude-Ret
& 55.86 & 0.896 & 0.957 & {0.899} & -1.71 & 11.1
& 45.99 & 0.870 & 0.958 & 0.884 & -1.27 & 16.1 \\
\textsc{SRAT}-Claude
& \textbf{61.10} & \textbf{0.898} & \textbf{0.961}
& \textbf{0.903} & \textbf{-1.62} & 50.3
& \textbf{49.03} & \textbf{0.875} & \textbf{0.967}
& \textbf{0.891} & \textbf{-1.04} & 25.1 \\
\bottomrule
\end{tabular}}
\caption{
Composition with a black-box translator. Claude-Ret uses prompted retrieval
decisions. \textsc{SRAT}-Claude keeps Claude frozen and uses the learned Qwen policy for retrieval and context construction.
}
\label{tab:blackbox}
\end{table*}

\subsection{RQ2: When Does Utility-Aligned Retriever Training Help?}
\label{sec:rq2}

We next examine whether the retriever can be better aligned with translation
utility. To valiate retrieval quality, we additionally calculate
{retrieval accuracy}: for each retrieval action, we ask Claude Sonnet 4.6 to judge whether the retrieved document is correct and useful for translating the source, given the query, source, reference, and retrieved document.

Retriever training is most beneficial when the underlying retrieval task is difficult. On IdiomKB, the default similarity retriever already achieves about 95\% retrieval accuracy, and joint or contrastive training improves it only to 97--98\%, yielding limited downstream gains. On RAGtrans, the default retriever reaches only 56.3\% and 72.6\% accuracy on En$\rightarrow$Zh and En$\rightarrow$De. Contrastive training increases these to 70.7\% and 81.4\%, while improving MQM by 6.8\% and 7.4\%. This shows that utility alignment is particularly valuable when semantic similarity is an insufficient proxy for evidence usefulness. We show such cases in Appendix \ref{app:similarity-cases}.

Contrastive training also consistently outperforms joint optimization. On RAGtrans, it improves MQM over joint training from $-4.41$ to $-4.27$ on
En$\rightarrow$Zh and from $-5.89$ to $-5.54$ on En$\rightarrow$De.
This supports our motivation in Section~\ref{sec:contrastive}: joint training shares one trajectory reward across many token-level decisions but only a few document selections, weakening retriever credit assignment. Contrastive training instead derives direct document-level supervision from controlled candidate interventions, providing a more focused learning signal. Additional
alternating-training curves are given in Appendix \ref{app:multi-round}.

\subsection{RQ3: Can Selective Retrieval Improve Strong Black-Box Translators?}
\label{sec:rq3}

We next examine whether selective retrieval can improve a stronger black-box
translator whose parameters are inaccessible. Specifically, we freeze Claude Sonnet 4.6 and
train a Qwen-based \method policy to decide \emph{when} and \emph{what} to
retrieve, using Claude's final translation quality as reward. We compare
against direct Claude translation and \textbf{Claude-Ret}, which prompts Claude itself to make retrieval decisions without policy training.

Table~\ref{tab:blackbox} shows that directly prompting a strong LLM to retrieve
is not consistently beneficial. On IdiomKB Zh$\rightarrow$En, Claude-Ret reduces COMET from 0.826 to 0.821 and MQM from $-0.92$ to $-0.99$; on RAGtrans En$\rightarrow$De, it similarly degrades BLEU, COMET, and MQM.
In contrast, \textsc{SRAT}-Claude provides more reliable gains. On IdiomKB, it improves most metrics over direct Claude with only 1.9--10.2\% retrieval, although the gains remain modest given Claude's strong parametric knowledge on the relatively simple idiom-translation tasks.

The benefit becomes substantially larger on the more challenging RAGtrans
benchmark. \textsc{SRAT}-Claude improves MQM over direct Claude by 12.4\% on En$\rightarrow$Zh and 16.8\% on En$\rightarrow$De, together with consistent gains in BLEU, COMET, and GRF. We further observe the same overall trend with GLM 5 in Appendix~\ref{app:glm}: prompted retrieval degrades translation,
whereas a trained \method policy can recover the loss and improve over
direct GLM translation. These results show that retrieval policy learning can be decoupled from the translator and generalized across frozen black-box LLMs, improving their translation outcomes without modifying their parameters.

\section{Conclusion}

We introduced Selective Retrieval-Augmented Translation (\method), which learns
\emph{when} and \emph{what} to retrieve for LLM-based translation. Across
offline and online retrieval settings, \method consistently outperforms fixed
never- and always-retrieve strategies. We further show that utility-aligned
retriever training improves difficult retrieval settings and that adaptive
retrieval can enhance a frozen black-box translator beyond direct translation.

\bibliography{references}

\clearpage

\appendix

\section{Experiment Details}\label{sec:expdetail}

\subsection{Results on Document-MT-LLM}\label{app:document-mt}

We show the evaluation results here on Document-MT-LLM \cite{wang2023document} , a document-level
En$\rightarrow$Ru subtitle translation task. Unlike IdiomKB and RAGtrans,
this setting has no predefined retrieval corpus. Instead, the policy needs to generate queries for a live search engine and incorporates the returned information into translation. This introduces additional challenges from both query generation and noisy, open-ended search results.

Table~\ref{tab:document-mt-full} shows that selective retrieval remains
effective beyond fixed offline corpora. \method achieves the best COMET, GRF, GRB, and MQM among all non-oracle systems, improving over
\textsc{SRAT-A} by 1.6\% COMET, 2.7\% GRF, and 5.4\% MQM.
The learned policy retrieves on 88.6\% of examples, substantially more often than on the sentence-level benchmarks, consistent with the greater contextual demands of long-form subtitle translation. Nevertheless, its advantage over always retrieving shows that even in this retrieval-intensive setting, external evidence is not uniformly beneficial. These results extend the partial-evidence observation to open-ended retrieval, where the policy must learn both whether additional context is needed and how to formulate an effective search query.

\begin{table}[h]
\centering
\scriptsize
\setlength{\tabcolsep}{2.0pt}
\renewcommand{\arraystretch}{0.92}
\resizebox{0.98\columnwidth}{!}{%
\begin{tabular}{lrrrrrr}
\toprule
Model & BLEU & COMET & GRF & GRB & MQM & Ret.\,(\%) \\
\midrule
Direct
& 17.01 & 0.690 & 0.601 & 0.544 & -20.78 & -- \\
KB-CoT
& 18.60 & 0.708 & 0.622 & 0.554 & -19.73 & 100.0 \\
Self-CoT
& 19.29 & 0.746 & 0.715 & 0.640 & -13.86 & -- \\
Ladder
& 24.96 & 0.796 & 0.831 & 0.740 & -6.87 & -- \\
CSC
& \textbf{25.01} & 0.800 & 0.829 & 0.737 & -7.06 & 100.0 \\
\midrule
\textsc{SRAT-N}
& 23.21 & 0.801 & 0.830 & 0.740 & -7.09 & 0.0 \\
\textsc{SRAT-A}
& 23.32 & 0.804 & 0.827 & 0.743 & -7.02 & 100.0 \\
\textsc{SRAT}
& 23.66 & \textbf{0.817} & \textbf{0.849} & \textbf{0.753}
& \textbf{-6.64} & 88.6 \\
\bottomrule
\end{tabular}}
\caption{
Results on Document-MT-LLM for En$\rightarrow$Ru translation with online
retrieval.
}
\label{tab:document-mt-full}
\end{table}

\subsection{Technical Details}\label{sec:expdetail}

We conduct all experiments on a cluster where each node is equipped with 8 A100 80GB GPU with 256 GB of RAM . For all models, we use the Adam optimizer and tune the learning rate within the range ([1e-7, 5e-7, 1e-6, 5e-6, 1e-5]). By default, we fine-tune
{Qwen2.5-7B-Instruct} using GRPO \citep{shaoetal2024deepseekmath}. GRPO is configured with a KL coefficient of 0.001 and a clip ratio of 0.2.
During rollout, we use a group size of 8 and a sampling temperature of 1.0. We enable tool-calling via {SGLang} \citep{zheng2024sglang} to allow the LLM to generate retrieval queries and set maximal turns to 5. For retrieval, we use the {Flat E5} retriever \citep{wang2022text} following prior work \citep{jin2025search}. Flat indexing encodes the corpus into dense embeddings and retrieves relevant documents via efficient embedding matching in the embedding space. If retrieval fails, the retriever returns a blank placeholder. On average, training the policy to convergence takes 68/47, 334/382, and 97 GPU hours on the IdiomKB (Zh2En/En2Zh), RAGtrans (En2Zh), and Document-MT-LLM (En2Ru) datasets, respectively. 

For all baselines, we follow the original papers for implementation details and hyperparameter tuning. We run each experiment with three random seeds and report the mean performance. Statistical significance  is evaluated using a paired t-test with $p < 0.05$.
All datasets used in our experiments are publicly available. All datasets used in our experiments are publicly available. They are either released under the MIT license, which permits research use, or come without a specific license, but their original papers explicitly state that public research usage is permitted. Our use of these datasets complies with their original license terms and respects the rights of their creators.

\subsection{Comparison with Oracle Retrieval Policy}
\label{app:oracle}

To estimate the remaining headroom in retrieval decisions, we construct an
\emph{Oracle} that selects, for each test example, the better output from
\textsc{SRAT-N} and \textsc{SRAT-A} according to the same blended quality
score used in RL reward. The Oracle therefore represents an idealized
per-instance choice between translating without retrieval and translating with retrieval, while keeping the underlying policies fixed.

As shown in Table \ref{tab:oracle}, the learned \method policy approaches the Oracle on several settings, while
the remaining gap reveals further room for improving retrieval decisions. On IdiomKB Zh$\rightarrow$En, for example, \method reaches 0.822 COMET and
$-2.74$ MQM compared with 0.831 and $-2.67$ for the Oracle. The largest
headroom appears on RAGtrans, particularly En$\rightarrow$De, where the
Oracle reaches 0.869 COMET and $-5.25$ MQM compared with 0.857 and $-5.98$
for \method. This is consistent with the greater difficulty of deciding when retrieval is useful in knowledge-intensive translation.

\begin{table*}[t]
\centering
\scriptsize
\setlength{\tabcolsep}{2.0pt}
\renewcommand{\arraystretch}{0.92}
\resizebox{0.86\linewidth}{!}{%
\begin{tabular}{lrrrrrrrrrrrr}
\toprule
& \multicolumn{6}{c}{En$\rightarrow$Zh}
& \multicolumn{6}{c}{En$\rightarrow$De} \\
\cmidrule(lr){2-7}
\cmidrule(lr){8-13}
Model
& BLEU & COMET & GRF & GRB & MQM & Ret.\,(\%)
& BLEU & COMET & GRF & GRB & MQM & Ret.\,(\%) \\
\midrule
GLM 5
& 58.58 & 0.892 & \textbf{0.953} & 0.891 & -1.666 & 0.0
& 46.35 & 0.869 & 0.956 & 0.882 & -1.668 & 0.0 \\
GLM-Ret
& 57.98 & 0.891 & 0.945 & 0.887 & -2.076 & 68.1
& 46.41 & 0.868 & 0.952 & 0.880 & -1.956 & 79.7 \\
\textsc{SRAT}-GLM
& \textbf{58.82} & \textbf{0.893} & \textbf{0.953}
& \textbf{0.892} & \textbf{-1.661} & 32.3
& \textbf{46.52} & \textbf{0.870} & \textbf{0.957}
& \textbf{0.883} & \textbf{-1.662} & 24.0 \\
\bottomrule
\end{tabular}}
\caption{
Results with GLM 5 as the frozen black-box translator on RAGtrans.
GLM-Ret uses prompted retrieval decisions, while \textsc{SRAT}-GLM trains
a separate Qwen policy for retrieval and context construction using GLM's
translation quality as reward.
}
\label{tab:glm}
\end{table*}

\begin{table}[t]
\centering
\fontsize{8.2pt}{9.2pt}\selectfont
\setlength{\tabcolsep}{1.7pt}
\renewcommand{\arraystretch}{1}
\begin{tabular*}{\columnwidth}{@{\extracolsep{\fill}}llrrrrrr@{}}
\toprule
Model & Dir. & BLEU & COMET & J-F & J-B & MQM & Ret. \\
\midrule
\multicolumn{8}{l}{{[IdiomKB]}} \\
SRAT
& Zh$\rightarrow$En
& 24.62 & 0.822 & 0.731 & 0.728 & -2.74 & 30.6 \\
Oracle
& Zh$\rightarrow$En
& 25.27 & 0.831 & 0.739 & 0.726 & -2.67 & 38.5 \\
\addlinespace[1pt]
SRAT
& En$\rightarrow$Zh
& 48.25 & 0.891 & 0.792 & 0.871 & -1.85 & 77.0 \\
Oracle
& En$\rightarrow$Zh
& 48.11 & 0.897 & 0.786 & 0.860 & -1.74 & 57.2 \\
\midrule
\multicolumn{8}{l}{{[RAGtrans]}} \\
SRAT
& En$\rightarrow$Zh
& 52.37 & 0.888 & 0.939 & 0.860 & -4.58 & 52.2 \\
Oracle
& En$\rightarrow$Zh
& 53.36 & 0.890 & 0.940 & 0.857 & -3.74 & 37.7 \\
\addlinespace[1pt]
SRAT
& En$\rightarrow$De
& 36.84 & 0.857 & 0.917 & 0.826 & -5.98 & 43.0 \\
Oracle
& En$\rightarrow$De
& 38.57 & 0.869 & 0.930 & 0.841 & -5.25 & 30.4 \\
\midrule
\multicolumn{8}{l}{{[Document-MT]}} \\
SRAT
& En$\rightarrow$Ru
& 23.66 & 0.817 & 0.849 & 0.753 & -6.64 & 88.6 \\
Oracle
& En$\rightarrow$Ru
& 24.87 & 0.821 & 0.851 & 0.758 & -6.52 & 85.0 \\
\bottomrule
\end{tabular*}
\caption{
Comparison between SRAT and the Oracle retrieval policy.
J-F/J-B correspond to I5/I3 on IdiomKB and GRF/GRB on RAGtrans and
Document-MT; Ret. denotes retrieval rate (\%). The Oracle selects between
\textsc{SRAT-N} and \textsc{SRAT-A} for each example using the quality score
in the RL reward.
}
\label{tab:oracle}
\end{table}

Interestingly, the Oracle often retrieves less frequently than \method,
especially on RAGtrans (37.7\% vs.\ 52.2\% for En$\rightarrow$Zh and
30.4\% vs.\ 43.0\% for En$\rightarrow$De). This suggests that further gains
may come not from increasing retrieval frequency, but from more accurately
identifying the subset of inputs for which retrieved evidence improves the
translation. Because the Oracle selects using the blended training score, it is an oracle for this selection objective rather than a strict upper bound on every individual evaluation metric.

\subsection{Learned Policy Behaviors}
\label{app:policy-behavior}

Beyond retrieval frequency, we further analyze how \method organizes its interaction
trajectory across tasks. Specifically, we measure \emph{multi-retrieval} ratio, the fraction of
examples for which more than one retrieval query is issued, and
\emph{reasoning} ratio, the fraction for which the model generates an explicit
reasoning block before translation.

Table~\ref{tab:policy-behavior} shows that the learned policy adapts not only
\emph{whether} to retrieve but also how much interaction and reasoning to use.
Multi-retrieval is rare on IdiomKB (0.6--1.4\%), increases on RAGtrans
(1.9--2.1\%), and is highest on Document-MT-LLM (3.1\%), suggesting that
harder or longer inputs occasionally require multiple queries to assemble
sufficient context.

Reasoning behavior varies even more strongly across tasks. It is almost absent
on IdiomKB (0--0.1\%), where retrieval often reduces to a direct idiom lookup,
but appears on 89--92\% of RAGtrans examples, where the model must frequently
assess terminology and factual knowledge before deciding whether retrieval is
necessary. Document-MT-LLM lies between these regimes at 66.3\%, reflecting a
mixture of self-contained dialogue and context-dependent segments. Together
with the task-dependent retrieval rates, these patterns indicate that \method
learns a task-conditioned interaction strategy rather than a fixed retrieve-
then-translate procedure.

\begin{table}[t]
\centering
\fontsize{8.2pt}{9.2pt}\selectfont
\setlength{\tabcolsep}{2.0pt}
\renewcommand{\arraystretch}{1.0}
\begin{tabular*}{0.94\columnwidth}{@{\extracolsep{\fill}}llrcc@{}}
\toprule
Dataset & Dir. & Ret. & Multi-Ret. & Reason. \\
\midrule
IdiomKB
& Zh$\rightarrow$En & 30.6 & 0.6 & 0.0 \\
& En$\rightarrow$Zh & 77.0 & 1.4 & 0.1 \\
\midrule
RAGtrans
& En$\rightarrow$Zh & 52.2 & 2.1 & 89.2 \\
& En$\rightarrow$De & 43.0 & 1.9 & 92.1 \\
\midrule
Document-MT
& En$\rightarrow$Ru & 88.6 & 3.1 & 66.3 \\
\bottomrule
\end{tabular*}
\caption{
Learned retrieval and reasoning behaviors of \method. Ret., Multi-Ret.,
and Reason. denote retrieval, multi-retrieval, and reasoning rates (\%),
respectively.
}
\label{tab:policy-behavior}
\end{table}

\subsection{Generalization to Another Black-Box Translator}
\label{app:glm}

To test whether the black-box adaptation of \method generalizes beyond Claude,
we further use GLM 5 as the frozen translator. We follow the same setup as
Section~\ref{sec:rq3}: \textbf{GLM} performs direct translation,
\textbf{GLM-Ret} prompts GLM itself to decide when and what to retrieve, and
\textbf{\textsc{SRAT}-GLM} trains a separate Qwen-based retrieval policy using
the quality of GLM's final translation as reward.

Table~\ref{tab:glm} shows a pattern consistent with the Claude experiments.
Directly prompting GLM to retrieve substantially hurts translation quality:
GLM-Ret degrades MQM from $-1.666$ to $-2.076$ on
En$\rightarrow$Zh and from $-1.668$ to $-1.956$ on En$\rightarrow$De,
while retrieving on 68.1\% and 79.7\% of inputs. In contrast,
\textsc{SRAT}-GLM reduces retrieval to 32.3\% and 24.0\%, respectively,
recovers the degradation, and slightly outperforms direct GLM across the
reported metrics. This further confirms that adaptive retrieval policy
learning can be decoupled from the translator and generalized across different
black-box LLMs.

The gains over direct GLM are nevertheless smaller than those observed with
Claude. Our qualitative analysis suggests that GLM is more likely to overuse
retrieved content rather than selectively incorporate the information relevant
to the current sentence, reducing the headroom from retrieval augmentation.
This highlights that the value of retrieved evidence depends not only on the
retrieval policy but also on how effectively the downstream translator can
exploit the resulting context. We provide representative comparisons in
Section~\ref{app:cases}.

\section{Additional Ablations and Robustness}
\label{app:robustness}

\begin{table}[t]
\centering
\scriptsize
\setlength{\tabcolsep}{2.3pt}
\renewcommand{\arraystretch}{0.94}
\resizebox{0.88\linewidth}{!}{%
\begin{tabular}{lrrrrrr}
\toprule
& \multicolumn{3}{c}{Opus 4.8}
& \multicolumn{3}{c}{DeepSeek 3.2} \\
\cmidrule(lr){2-4}
\cmidrule(lr){5-7}
Model & I5 & I3 & MQM & I5 & I3 & MQM \\
\midrule
Direct
& 0.704 & 0.711 & -3.52 & 0.464 & 0.379 & -1.94 \\
KB-CoT
& 0.736 & 0.756 & -3.12 & 0.498 & 0.412 & -1.64 \\
Self-CoT
& 0.726 & 0.754 & -3.10 & 0.481 & 0.402 & -1.97 \\
Ladder
& 0.829 & 0.857 & -1.47 & 0.545 & 0.450 & -1.25 \\
CSC
& 0.840 & 0.874 & -1.37 & 0.558 & 0.464 & -1.33 \\
\textsc{SRAT-N}
& 0.833 & 0.870 & -1.42 & 0.543 & 0.458 & -1.29 \\
\textsc{SRAT-A}
& 0.837 & 0.878 & -1.36 & 0.549 & 0.460 & -1.24 \\
\textsc{SRAT}
& \textbf{0.862} & \textbf{0.907} & \textbf{-1.20}
& \textbf{0.587} & \textbf{0.484} & \textbf{-0.90} \\
\bottomrule
\end{tabular}}
\caption{
IdiomKB En$\rightarrow$Zh results under alternative LLM evaluators.
}
\label{tab:judge-idiom}
\end{table}

\begin{table}[t]
\centering
\scriptsize
\setlength{\tabcolsep}{2.3pt}
\renewcommand{\arraystretch}{0.94}
\resizebox{0.88\linewidth}{!}{%
\begin{tabular}{lrrrrrr}
\toprule
& \multicolumn{3}{c}{Opus 4.8}
& \multicolumn{3}{c}{DeepSeek 3.2} \\
\cmidrule(lr){2-4}
\cmidrule(lr){5-7}
Model & GRF & GRB & MQM & GRF & GRB & MQM \\
\midrule
Direct
& 0.707 & 0.713 & -13.95 & 0.831 & 0.804 & -10.35 \\
KB-CoT
& 0.711 & 0.715 & -13.69 & 0.836 & 0.815 & -10.37 \\
Self-CoT
& 0.734 & 0.735 & -12.44 & 0.848 & 0.825 & -9.76 \\
Ladder
& 0.874 & 0.855 & -4.65 & 0.924 & 0.890 & -3.01 \\
CSC
& 0.872 & 0.850 & -4.80 & 0.920 & 0.891 & -3.10 \\
\textsc{SRAT-N}
& 0.870 & 0.852 & -4.59 & 0.922 & 0.890 & -3.05 \\
\textsc{SRAT-A}
& 0.877 & 0.849 & -4.73 & 0.926 & 0.889 & -2.92 \\
\textsc{SRAT}
& \textbf{0.889} & \textbf{0.869} & \textbf{-4.12}
& \textbf{0.939} & \textbf{0.906} & \textbf{-2.57} \\
\bottomrule
\end{tabular}}
\caption{
RAGtrans En$\rightarrow$De results under alternative LLM evaluators.
}
\label{tab:judge-rag}
\end{table}

We conduct additional studies to verify that the improvements of \method are
not specific to a particular training stage, optimization schedule, backbone,
LLM evaluator, or retrieval backend.

\subsection{Contribution of SFT and RL}
\label{app:sft-rl}

\method uses SFT on high-quality retrieval trajectories as initialization,
followed by outcome-driven RL. To separate their contributions, we compare the
intermediate SFT checkpoint with the final model on representative RAGtrans and
IdiomKB settings.

\begin{table}[t]
\centering
\fontsize{8.0pt}{9.0pt}\selectfont
\setlength{\tabcolsep}{2.4pt}
\renewcommand{\arraystretch}{1.0}
\begin{tabular*}{\columnwidth}{@{\extracolsep{\fill}}llrrrrr@{}}
\toprule
Task & Model & BLEU & COMET & J-F & J-B & MQM \\
\midrule
\multirow{3}{*}{En$\rightarrow$Zh}
& CSC
& \textbf{55.42} & 0.880 & 0.920 & 0.840 & -5.40 \\
& SRAT (SFT)
& 55.34 & 0.875 & 0.918 & 0.837 & -5.45 \\
& SRAT (SFT+RL)
& 52.37 & \textbf{0.888} & \textbf{0.939} & \textbf{0.860}
& \textbf{-4.58} \\
\midrule
\multirow{3}{*}{Zh$\rightarrow$En}
& CSC
& \textbf{24.95} & 0.812 & 0.635 & 0.643 & -4.01 \\
& SRAT (SFT)
& 24.91 & 0.815 & 0.640 & 0.644 & -3.95 \\
& SRAT (SFT+RL)
& 24.62 & \textbf{0.822} & \textbf{0.731} & \textbf{0.728}
& \textbf{-2.74} \\
\bottomrule
\end{tabular*}
\caption{
Contribution of SFT and RL. En$\rightarrow$Zh uses RAGtrans, where
J-F/J-B denote GRF/GRB; Zh$\rightarrow$En uses IdiomKB, where they denote
I5/I3.
}
\label{tab:sft-rl}
\end{table}

Table~\ref{tab:sft-rl} shows that the SFT warm start already provides a strong
initialization, performing similarly to CSC. The main quality improvements,
however, emerge after RL. On IdiomKB Zh$\rightarrow$En, for example, I5
increases from 0.640 to 0.731 and MQM improves from $-3.95$ to $-2.74$.
BLEU decreases slightly after RL, consistent with Section~\ref{sec:rq1}:
the RL reward directly optimizes COMET and LLM-judged quality rather than
reference n-gram overlap.

\begin{table*}[t]
\centering
\scriptsize
\setlength{\tabcolsep}{2.0pt}
\renewcommand{\arraystretch}{0.94}
\resizebox{0.92\linewidth}{!}{%
\begin{tabular}{lrrrrrrlrrrrrr}
\toprule
& \multicolumn{6}{c}{RAGtrans: En$\rightarrow$De}
& \multicolumn{6}{c}{IdiomKB: En$\rightarrow$Zh} \\
\cmidrule(lr){2-7}
\cmidrule(lr){8-13}
Model
& BLEU & COMET & GRF & GRB & MQM & Ret.
& BLEU & COMET & I5 & I3 & MQM & Ret. \\
\midrule
Direct
& 34.82 & 0.8481 & 0.8714 & 0.7898 & -9.37 & --
& 45.99 & 0.8687 & 0.665 & 0.716 & -2.96 & -- \\
KB-CoT
& 34.56 & 0.8453 & 0.8724 & 0.7856 & -9.09 & 100.0
& 46.52 & 0.8728 & 0.691 & 0.743 & -2.57 & 100.0 \\
Self-CoT
& 33.51 & 0.8456 & 0.8692 & 0.7883 & -8.89 & --
& 46.03 & 0.8674 & 0.676 & 0.730 & -2.88 & -- \\
Ladder
& 42.11 & 0.8592 & 0.9239 & 0.8384 & -4.39 & --
& 54.04 & 0.8890 & 0.754 & 0.822 & -2.00 & -- \\
CSC
& \textbf{42.28} & 0.8611 & 0.9229 & 0.8406 & -4.20 & 100.0
& \textbf{54.27} & 0.8902 & 0.767 & 0.829 & -1.87 & 100.0 \\
\midrule
\textsc{SRAT-N}
& 39.47 & 0.8597 & 0.9230 & 0.8398 & -4.21 & --
& 52.94 & 0.8898 & 0.763 & 0.824 & -1.92 & -- \\
\textsc{SRAT-A}
& 40.26 & 0.8613 & 0.9235 & 0.8427 & -4.07 & 100.0
& 53.27 & 0.8991 & 0.770 & 0.836 & -1.65 & 100.0 \\
\textsc{SRAT}
& 41.37 & \textbf{0.8720} & \textbf{0.9318} & \textbf{0.8605}
& \textbf{-3.88} & 37.6
& 53.94 & \textbf{0.9112} & \textbf{0.789} & \textbf{0.858}
& \textbf{-1.33} & 79.1 \\
\bottomrule
\end{tabular}}
\caption{
Results with Qwen3-8B. Ret. denotes retrieval rate (\%).
}
\label{tab:qwen3}
\end{table*}

\begin{table*}[t]
\centering
\scriptsize
\setlength{\tabcolsep}{2.2pt}
\renewcommand{\arraystretch}{0.96}
\resizebox{0.88\linewidth}{!}{%
\begin{tabular}{lrrrrrr rrrrrr}
\toprule
& \multicolumn{6}{c}{IdiomKB: Zh$\rightarrow$En}
& \multicolumn{6}{c}{RAGtrans: En$\rightarrow$De} \\
\cmidrule(lr){2-7}
\cmidrule(lr){8-13}
Top-$k$
& BLEU & COMET & I5 & I3 & MQM & Ret.
& BLEU & COMET & GRF & GRB & MQM & Ret. \\
\midrule
$k=1$
& 24.62 & \textbf{0.822} & \textbf{0.731} & \textbf{0.728}
& \textbf{-2.74} & 30.6
& 36.84 & 0.857 & \textbf{0.917} & \textbf{0.826}
& \textbf{-5.98} & 43.3 \\

$k=3$
& \textbf{24.68} & 0.817 & 0.729 & 0.721
& -2.88 & 27.1
& \textbf{36.88} & \textbf{0.860} & 0.913 & 0.820
& -6.08 & 40.1 \\

$k=5$
& 23.11 & 0.804 & 0.713 & 0.708
& -3.13 & 22.4
& {34.71} & {0.795} & 0.880 & 0.803
& -9.44 & 31.7 \\
\bottomrule
\end{tabular}}
\caption{
Effect of the number of documents returned per retrieval action.
Ret. denotes the percentage of examples with at least one retrieval call.
Higher is better for all translation metrics, including MQM where values
closer to zero are better.
}
\label{tab:topk}
\end{table*}

\subsection{Single-Round vs.\ Multi-Round Optimization}
\label{app:multi-round}

Our contrastive retriever training can naturally be alternated with translator
RL for multiple rounds. We therefore evaluate two rounds on RAGtrans:
\textbf{Direct} denotes the initial model, $\mathrm{T}_1/\mathrm{T}_2$
update the translator with the retriever frozen, and
$\mathrm{R}_1/\mathrm{R}_2$ update the retriever with the translator frozen.

\begin{figure}[t]
    \centering
    \includegraphics[width=1.02\columnwidth]{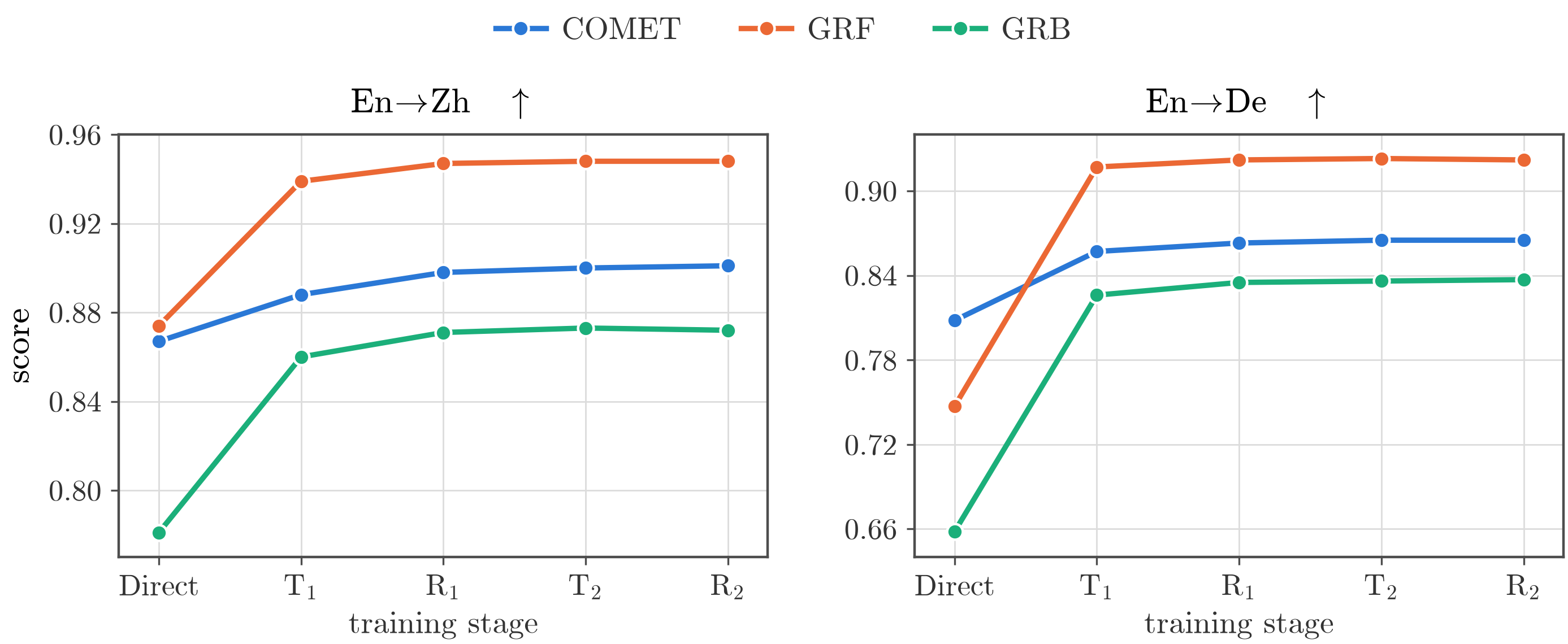}
    \caption{
    Alternating translator and retriever optimization on RAGtrans.
    $\mathrm{T}_1/\mathrm{T}_2$ denote translator RL updates and
    $\mathrm{R}_1/\mathrm{R}_2$ denote contrastive retriever updates.
    }
    \label{fig:multi-round}
\end{figure}

Figure~\ref{fig:multi-round} shows clear diminishing returns. On
En$\rightarrow$De, the first round improves GRF/GRB by 0.175/0.177, whereas
the second adds only 0.000/0.002; on En$\rightarrow$Zh, the corresponding GRF
gains are 0.073 and 0.001. Most of the first-round improvement comes from
translator RL, with retriever refinement providing some additional gains.
Since another round requires both a full RL stage and retriever training for
negligible improvement, we use a single round by default.

\subsection{Robustness to LLM Evaluators}
\label{app:judges}

Our main evaluation uses Claude Sonnet 4.6. To test whether the observed gains
depend on a particular judge, we re-evaluate all systems using Claude Opus 4.8 and DeepSeek 3.2. Results are shown in Table \ref{tab:judge-rag} and Table \ref{tab:judge-idiom}.

Across both datasets and both alternative judges, \method achieves the best
score on every reported metric. More importantly, the relative system ordering
is highly consistent with the Sonnet-based evaluation in the main experiments.
The gains therefore persist across substantially different evaluator models,
indicating that they are not artifacts of optimizing toward one particular
LLM judge.

\subsection{Robustness to a Stronger Backbone}
\label{app:qwen3}

We replace Qwen2.5-7B-Instruct with Qwen3-8B to test whether selective
retrieval remains effective with a stronger backbone. We evaluate RAGtrans
En$\rightarrow$De and IdiomKB En$\rightarrow$Zh and report results in Table \ref{tab:qwen3}.

The results reproduce the main pattern with a stronger backbone: \method
achieves the best COMET, task-specific judge, and MQM scores in both settings
while retrieving selectively. On RAGtrans, MQM improves by 7.6\% over CSC
and 4.7\% over \textsc{SRAT-A}; on IdiomKB, it improves MQM from $-1.87$
for CSC to $-1.33$. Thus, the benefit of selective retrieval is not specific
to the Qwen2.5-7B backbone.

\begin{table*}[t]
\centering
\small
\begin{tabular}{lcccccc}
\toprule
Method & Accuracy & Fluency & Style & Terminology & Non-translation & Total MQM \\
\midrule
Direct & -1.73 & -1.55 & -0.92 & -1.61 & -0.01 & -6.32 \\
CSC    & -1.30 & -0.60 & -0.86 & -0.78 & -0.02 & -3.56 \\
SRAT-A & -1.15 & -0.58 & -0.73 & -0.31 & \textbf{0.00} & -2.77 \\
SRAT   & \textbf{-0.81} & \textbf{-0.39} & \textbf{-0.50}
       & \textbf{-0.24} & \textbf{0.00} & \textbf{-1.94} \\
\bottomrule
\end{tabular}
\caption{Human evaluation on 32 En$\rightarrow$Zh test examples from IdiomKB and RAGtrans. Scores are averaged across examples and annotators. Higher is better for all dimensions and Total MQM.}
\label{tab:human-evaluation}
\end{table*}

\subsection{Effect of the Number of Retrieved Documents}
\label{app:topk}

By default, \method retrieves only the top-$1$ document for each query. We
further vary the number of returned documents to $k\in\{1,3,5\}$ to examine
whether exposing the translator to more candidate evidence improves downstream
translation. The policy and training procedure are otherwise unchanged.

Table~\ref{tab:topk} shows that retrieving more documents does not consistently
improve translation quality. On IdiomKB, $k=1$ achieves the strongest COMET,
I5, I3, and MQM, while increasing to $k=5$ substantially degrades all metrics.
On RAGtrans, $k=3$ slightly improves BLEU and COMET, but reduces GRF, GRB, and
MQM relative to $k=1$; increasing further to $k=5$ also decreases the performance. Retrieval rates also decrease as more documents are returned per action.

These results suggest that retrieval quality is more important than retrieval
quantity. Additional top-ranked documents can introduce redundant,
distracting, or even harmful context without providing greater translation utility, consistent
with the qualitative examples in Section~\ref{app:similarity-cases}, where
semantic similarity alone does not reliably identify the evidence most useful
for translation. We therefore use $k=1$ as the default throughout the paper.

\subsection{Robustness to Online Retrieval Backends}
\label{app:search-backend}

Finally, we test whether the online-retrieval results depend on a particular search engine. Keeping the learned policy fixed, we replace the default DuckDuckGo backend with Google Search on Document-MT-LLM.

Table~\ref{tab:search-backend} shows that replacing DuckDuckGo with Google
improves every translation metric while leaving the policy unchanged. This suggests that \method can exploit higher-quality retrieved evidence rather than depending on a particular search backend, and that improvements to the
underlying retrieval service can translate directly into stronger downstream translation quality.

\begin{table}[h]
\centering
\fontsize{8.0pt}{9.0pt}\selectfont
\setlength{\tabcolsep}{2.2pt}
\renewcommand{\arraystretch}{1.0}
\begin{tabular*}{\columnwidth}{@{\extracolsep{\fill}}lrrrrrr@{}}
\toprule
Retriever & BLEU & COMET & GRF & GRB & MQM & Ret. \\
\midrule
DuckDuckGo
& 23.66 & 0.817 & 0.849 & 0.753 & -6.64 & 88.6 \\
Google
& \textbf{23.98} & \textbf{0.826} & \textbf{0.856}
& \textbf{0.761} & \textbf{-6.53} & 91.3 \\
\bottomrule
\end{tabular*}
\caption{
Online retrieval backends on Document-MT-LLM En$\rightarrow$Ru.
Ret. denotes retrieval rate (\%).
}
\label{tab:search-backend}
\end{table}

\subsection{Human Evaluation}
\label{app:human-evaluation}

Finally, to assess whether the improvements observed under automatic evaluation are reflected in human judgments, we conduct a blinded human evaluation on En$\rightarrow$Zh translation. We randomly select 32 test examples, with 16 each from IdiomKB and RAGtrans, and collect translations from four systems: Direct, CSC, SRAT-A, and SRAT. We recruit 20 native Chinese speakers with advanced English proficiency as annotators. They independently evaluate the translations with system identities hidden and system outputs randomly assigned to columns. Following the MQM criteria in Appendix~\ref{app:judge-prompts}, annotators assign error penalties across five dimensions: Accuracy, Fluency, Style, Terminology, and Non-translation. Scores are reported as negative penalties, with values closer to zero indicating better quality.

Table~\ref{tab:human-evaluation} shows that human judgments support the overall ranking observed under automatic evaluation: SRAT ($-1.94$) outperforms SRAT-A ($-2.77$), followed by CSC ($-3.56$) and Direct ($-6.32$). Compared with Direct, SRAT reduces accuracy and terminology penalties by 53.2\% and 85.1\%, respectively, while also improving fluency and style. Relative to SRAT-A, SRAT achieves better scores across all four dimensions, with the largest absolute improvement in accuracy, from $-1.15$ to $-0.81$. Non-translation penalties are negligible across systems. These results provide additional evidence that selective retrieval improves translation quality beyond an always-retrieve strategy, with benefits extending beyond terminology to overall accuracy and expression.

Last, we provide more details on how the human evaluation was conducted. All 20 annotators were native Chinese speakers enrolled in master's or PhD programs at a U.S. university, with advanced English proficiency. They were recruited through a shared link to an online annotation form. Participation was entirely voluntary and unpaid, and annotators provided informed consent through the form both to participate and to have their annotations collected and used for the research reported in this paper. Annotators received the same MQM evaluation instructions reproduced in Appendix~\ref{app:judge-prompts}. The annotation protocol was determined exempt from further ethics review by the relevant institutional ethics review board.

\section{Limitations}

Our evaluation covers several translation settings and language directions, while extending SRAT to a broader range of languages and specialized domains would further establish its applicability. Our human evaluation provides complementary evidence to automatic metrics, but remains relatively limited in scale. In addition, retriever alignment is learned for a particular translator, making the transfer of retrieval policies and evidence preferences across models a valuable direction. Finally, exploring SRAT on tasks that require combining evidence from more sources could further reveal the potential of sequential retrieval for translation on more complex and general scenarios.

\section{Ethical Considerations}

SRAT aims to improve translation quality, but retrieved sources may contain inaccurate or biased information that influences the resulting translations. Careful source selection and human review remain important, particularly for sensitive or high-stakes content. When using online retrieval, queries may expose information from the source text to external services, motivating privacy-aware deployment for confidential material. Translation quality may also vary across languages and communities, warranting broader evaluation before deployment. Finally, policy training and external model evaluations consume computational resources; reusing trained components and exploring more economical adaptation strategies could help reduce their environmental impact.

\section{Qualitative Case Studies}
\label{app:cases}

We provide representative cases illustrating four aspects of selective
retrieval: when external evidence helps, when retrieval can instead introduce
errors, why BLEU can disagree with semantic translation quality, and how
different black-box translators use the same retrieved evidence.

\begin{CJK*}{UTF8}{gbsn}

\subsection{When Retrieval Helps}
\label{app:cases-help}

The first group illustrates inputs for which parametric knowledge is
insufficient, while the retrieved evidence resolves an idiom,
named entity, or technical concept to help improve translation quality.

\begin{casebox}{Case 1: Resolving an Idiom in Context}

\casefield{Source}{
\trad{她第二次與活地·亞倫合作，在他的1999年電影《甜蜜與卑微》中飾演一個次要角色，
《Greenwich Village Gazette》稱其為「引人注目」。}
}

\casefield{Reference}{
For her second film with Woody Allen, 1999's \textit{Sweet and Lowdown},
she played a minor role which the Greenwich Village Gazette called
``notable''.
}

\casebar{\textsc{SRAT-N} --- No Retrieval}

She collaborated with Liv Ullmann for the second time, playing a minor role
in his 1999 film \textit{Sweet and Lowdown}, which was described by the
Greenwich Village Gazette as ``attractive''.

\casescore{
BLEU 19.21 \quad COMET 0.816 \quad I5 4.0 \quad MQM $-6.1$
}

\casebar{\textsc{SRAT} --- Retrieval}

She collaborated with Liv Tyler again in his 1999 film
\textit{Sweet and Lowdown}, playing a minor role described by the
Greenwich Village Gazette as ``remarkable''.

\casefield{Retrieved Evidence}{
Query: ``引人注目''.
Doc: ``attract attention; eye-catching''.
}

\casescore{
BLEU 22.42 \quad COMET 0.853 \quad I5 5.0 \quad MQM $-1.1$
}

\medskip

\casefield{Analysis}{
The direct output interprets ``引人注目'' as ``attractive'', which is
inappropriate in this context. The retrieved gloss clarifies the intended
sense and leads to ``remarkable'', close to the reference ``notable''.
The gain is localized to the idiom covered by the evidence.
}

\end{casebox}

\begin{casebox}{Case 2: Disambiguating an Abstract Idiom}

\casefield{Source}{
但实际上，听者可以从语义上推断出一个言外之意 (indirect speech act)，
即 ``Please pass the mustard''（请把芥末递给我）。
}

\casefield{Reference}{
There is, however, an indirect speech act that can optionally be inferred
from the question, namely ``please pass the mustard''.
}

\casebar{\textsc{SRAT-N} --- No Retrieval}

In fact, the listener can infer a ``presupposed meaning'' (indirect speech
act), such as ``Please pass the mustard''.

\casescore{
BLEU 15.40 \quad COMET 0.705 \quad I5 4.0 \quad MQM $-12.0$
}

\casebar{\textsc{SRAT} --- Retrieval}

In fact, the listener can infer an ``implied meaning'' (indirect speech act)
from the context, such as ``Please pass the mustard''.

\casefield{Retrieved Evidence}{
Query: ``言外之意''.
Doc: ``the unspoken meaning; what is implied but not explicitly stated''.
}

\casescore{
BLEU 16.93 \quad COMET 0.745 \quad I5 5.0 \quad MQM $-2.0$
}

\medskip

\casefield{Analysis}{
``Presupposed meaning'' is a related but distinct concept in pragmatics.
The retrieved definition disambiguates the expression and provides the
intended notion of implied meaning.
}

\end{casebox}

\begin{casebox}{Case 6: Grounding a Proper Name and Noble Rank}

\casefield{Source}{
Jean-Frédéric Phélypeaux, 1st Count of Maurepas was a French statesman
and Count of Maurepas.
}

\casefield{Reference}{
让-弗雷德里克·菲利波，第一代莫尔帕伯爵是法国政治家和莫尔帕伯爵。
}

\casebar{\textsc{SRAT-N} --- No Retrieval}

让-弗雷德里克·菲勒佩奥，第1代马尔帕斯子爵，是一位法国政治家，
并且是马尔帕斯子爵。

\casescore{
BLEU 33.36 \quad COMET 0.851 \quad GRF 0.882 \quad MQM $-15.1$
}

\casebar{\textsc{SRAT} --- Retrieval}

让-弗雷德里克·菲利波，第一代莫尔帕伯爵，是一位法国政治家及莫尔帕伯爵。

\casefield{Retrieved Evidence}{
Query: ``让-弗雷德里克·费利佩阿''.
Doc: ``【让-弗雷德里克·菲利波 (莫尔帕伯爵)】莫尔帕伯爵出生于凡尔赛一个行政贵族家庭,父亲蓬查特兰伯爵热罗姆·菲利波(Jérôme Phélypeaux),是海军和王室国务卿 ...''.
}

\casescore{
BLEU 76.84 \quad COMET 0.920 \quad GRF 0.948 \quad MQM $-2.0$
}

\medskip

\casefield{Analysis}{
Without evidence, the model approximates the surname phonetically and
mistranslates Count as Viscount. Retrieval supplies both the canonical
Chinese name and the correct noble rank.
}

\end{casebox}

\begin{casebox}{Case 7: Correcting a Technical Factual Error}

\casefield{Source}{
Winchester rifle is a comprehensive term describing a series of
lever-action repeating rifles manufactured by the Winchester Repeating
Arms Company.
}

\casefield{Reference}{
温彻斯特步枪是温彻斯特连发枪械公司生产的一系列杠杆式连发步枪的统称。
}

\casebar{\textsc{SRAT-N} --- No Retrieval}

温彻斯特步枪是一个综合术语，用于描述温彻斯特枪支公司制造的一系列
拉栓式连发步枪。

\casescore{
BLEU 45.19 \quad COMET 0.853 \quad GRF 0.895 \quad MQM $-25.0$
}

\casebar{\textsc{SRAT} --- Retrieval}

温彻斯特步枪是一个综合术语，用于描述温彻斯特连发武器公司制造的一系列
杠杆式连发步枪。

\casefield{Retrieved Evidence}{
Query: ``温彻斯特步枪''.
Doc: ``【温彻斯特步枪】1848年,美国人沃尔特·亨特开发出一种采用两个杠杆和复杂的接驳装置运作、发射被称为“火箭球”的金属弹丸,以内置式管状弹仓供弹的“意志连珠步枪 ...''.
}

\casescore{
BLEU 63.13 \quad COMET 0.884 \quad GRF 0.923 \quad MQM $-1.0$
}

\medskip

\casefield{Analysis}{
The direct model confuses lever-action with bolt-action (``拉栓式''),
a fluent but factual terminology error. The retrieved article provides the
correct mechanism.
}

\end{casebox}

\subsection{When Retrieval Hurts}
\label{app:cases-hurt}

The next cases illustrate why retrieval should not be applied uniformly.
Even relevant evidence may override a better direct translation, while
irrelevant or misleading evidence can introduce new errors.

\begin{casebox}{Case 3: A Literal Explanation Induces Literal Translation}

\casefield{Source}{
你别狗咬吕洞宾，不识好人心，我是真心帮你的。
}

\casefield{Reference}{
Don't bite the hand that feeds you---I'm genuinely trying to help you.
}

\casebar{\textsc{SRAT} --- Direct Translation}

Don't bite the hand that helps you; you don't recognize kindness, but I am
genuinely trying to help you.

\casescore{
BLEU 38.31 \quad COMET 0.715 \quad I5 5.0 \quad MQM $-0.1$
}

\casebar{\textsc{SRAT-A} --- Forced Retrieval}

You're just like the dog that bit Lü Dongbin; you don't know the value of
kindness. I was genuinely trying to help you.

\casefield{Retrieved Evidence}{
Query: ``狗咬吕洞宾''.
Doc: ``a dog biting Lü Dongbin ... meaning to fail to recognize good intentions''.
}

\casescore{
BLEU 19.70 \quad COMET 0.700 \quad I5 4.0 \quad MQM $-12.1$
}

\medskip

\casefield{Analysis}{
The retrieved explanation is correct, but it encourages literal transfer of
a culture-specific image into English. The direct model already produces a
natural English idiomatic equivalent.
}

\end{casebox}

\begin{casebox}{Case 4: Direct Translation Already Suffices}

\casefield{Source}{
不要火上浇油
}

\casefield{Reference}{
Don't fight fire with fire.
}

\casebar{\textsc{SRAT} --- Direct Translation}

Don't add fuel to the fire.

\casescore{
BLEU 15.68 \quad COMET 0.746 \quad I5 5.0 \quad MQM $-1.0$
}

\casebar{\textsc{SRAT-A} --- Forced Retrieval}

Don't make it worse.

\casefield{Retrieved Evidence}{
Query: ``不要火上浇油''.
Doc: ``add fuel to the fire; worsen the situation''.
}

\casescore{
BLEU 9.65 \quad COMET 0.5028 \quad I5 4.0 \quad MQM $-5.0$
}

\medskip

\casefield{Analysis}{
The model already knows the conventional English expression. Retrieval
instead pushes it toward a generic paraphrase and removes the figurative
register, despite slightly increasing sentence BLEU.
}

\end{casebox}

\begin{casebox}{Case 5: Query--Document Mismatch}

\casefield{Source}{
在生物学领域，吸附水和自由水的生物学应用也迥然不同——物理吸附水很难从
生物材料中被分离。
}

\casefield{Reference}{
In biological applications there can also be a distinction between
physisorbed water and "free" water — the physisorbed water being that closely associated with and relatively difficult to remove from a biological material.
}

\casebar{\textsc{SRAT} --- Direct Translation}

In the field of biology, adsorbed water and free water have markedly
different biological applications---physically adsorbed water is difficult
to separate from biological materials.

\casescore{
BLEU 10.37 \quad COMET 0.732 \quad I5 5.0 \quad MQM $-3.0$
}

\casebar{\textsc{SRAT-A} --- Forced Retrieval}

In biology, the biological applications of adsorbed water and free water
are very different---physical adsorbed water is attaching itself to the
biological materials and is difficult to separate from them.

\casefield{Retrieved Evidence}{
Query: ``吸附水''.
Doc: ``趋炎附势''---``to seek favor by attaching oneself to
the influential and powerful''.
}

\casescore{
BLEU 6.57 \quad COMET 0.614 \quad I5 4.0 \quad MQM $-13.0$
}

\medskip

\casefield{Analysis}{
The technical term ``吸附水'' is matched to an unrelated idiom sharing the
character ``附''. The retrieved notion of ``attaching oneself'' then leaks
into the translation. This case directly motivates utility-aligned retriever
training.
}

\end{casebox}

\begin{casebox}{Case 8: Retrieved Evidence Introduces Semantic Distortion}

\casefield{Source}{
An automated external defibrillator is a portable electronic device that
diagnoses life-threatening cardiac arrhythmias and treats them through
defibrillation, allowing the heart to re-establish an effective rhythm.
}

\casefield{Reference}{
自动体外心脏去颤器是一种便携式电子装置，能够自动诊断危及生命的室颤和
无脉性室性心动过速等心律失常，并能够通过除颤来进行治疗，使心脏重新建立
有效的节律。
}

\casebar{\textsc{SRAT} --- Direct Translation}

自动体外除颤器是一种便携式电子设备，能够自动诊断危及生命的室性心律失常
（包括室颤和无脉性室速），并通过除颤来治疗这些状况，使心脏能够恢复有效的节律。

\casescore{
BLEU 48.50 \quad COMET 0.887 \quad GRF 0.937 \quad MQM $-3.0$
}

\casebar{\textsc{SRAT-A} --- Forced Retrieval}

自动体外除颤器（AED）是一种便携式电子设备，能够自动诊断危及生命的室颤和
无脉性室速，并通过除颤来治疗这些心律失常，使心脏停止异常跳动，从而恢复
有效的心跳节奏。

\casefield{Retrieved Evidence}{
Query: ``自动体外除颤器''.
Doc: ``【自动体外心脏去颤器】自动体外心脏去颤器,于伤者脉搏停止时使用。然而它并不会对无心律,即心电图呈水平直线的伤者作出电击。 简而言之,使用去颤器并非让伤患者恢复心跳,那是许多节目的误导,而是使正在处于颤抖或是不正常跳动的心脏经由电流刺激而归零停止,并于电击完后立刻配合CPR以使不正常心跳的情况恢复正常 ...''.
}

\casescore{
BLEU 39.21 \quad COMET 0.852 \quad GRF 0.902 \quad MQM $-15.0$
}

\medskip

\casefield{Analysis}{
The retrieved description induces the erroneous claim that defibrillation
stops the heart's abnormal beating rather than the arrhythmia. Thus, even
topically relevant evidence can introduce factual distortion.
}

\end{casebox}

\begin{casebox}{Case 9: Wrong Entity Retrieved}

\casefield{Source}{
Geoffrey V, called the Handsome, the Fair or Plantagenet, was the Count of
Anjou, Touraine and Maine by inheritance from 1129, and also Duke of
Normandy by conquest from 1144.
}

\casefield{Reference}{
杰弗里五世，被称为英俊者、英美者或金雀花，从1129年起继承安茹、图赖讷和
缅因伯爵的头衔，并于1144年通过征服成为诺曼底公爵。
}

\casebar{\textsc{SRAT} --- Direct Translation}

杰弗里五世，又称英俊者、英明者或金雀花者，1129年因继承而成为安茹、
图赖讷和缅因伯爵，并于1144年因征服而成为诺曼底公爵。

\casescore{
BLEU 50.34 \quad COMET 0.940 \quad GRF 0.954 \quad MQM $-1.0$
}

\casebar{\textsc{SRAT-A} --- Forced Retrieval}

杰弗里五世（又称英俊者、英明者或普兰坦内特）继承于1129年成为安茹、
图尔奈和曼恩伯爵，并于1144年通过征服成为诺曼底公爵。

\casefield{Retrieved Evidence}{
Query: ``杰弗里五世''.
Doc: ``【詹姆斯五世 (苏格兰)】1512年4月10日,詹姆斯五世在林利思哥宫出生。他的兄长罗撒西公爵詹姆斯·斯图亚特及亚瑟·斯图亚特皆早夭,所以詹姆斯五世是詹姆斯四世唯一的子嗣,出生后翌日即被封为罗撒西公爵及苏格兰王子 ...''.
}

\casescore{
BLEU 44.33 \quad COMET 0.929 \quad GRF 0.941 \quad MQM $-25.0$
}

\medskip

\casefield{Analysis}{
The query is reasonable, but the retriever confuses two entities sharing
the ordinal ``五世''. Conditioning on the wrong biography corrupts proper
names that the model had translated correctly without retrieval.
}

\end{casebox}

\subsection{Semantic Similarity Does Not Imply Translation Utility}
\label{app:similarity-cases}

The following cases further illustrate why semantic similarity alone is an
imperfect criterion for retrieval. A document can strongly match the query
lexically while discussing only a narrower or tangential topic. Conversely,
a slightly lower-ranked document may provide the information actually needed
for translation.

\begin{casebox}{Case 14: Lexical Match versus Topical Utility}

\casefield{Source}{
Farming in North Korea is concentrated in the flatlands of the four west
coast provinces, where a longer growing season, level land, adequate rainfall,
and good irrigated soil permit the most intensive cultivation of crops. A
narrow strip of similarly fertile land runs through the eastern seaboard
Hamgyŏng provinces and Kangwŏn Province.
}

\casefield{Query}{
``朝鲜农业''
}

\casebar{Rank 1 --- Higher Semantic Similarity}

\textbf{Similarity:} 0.860120

\vspace{0.5mm}
朝鲜的农业自1948年建国后一直主要依靠由苏联提供的特惠粮食。然而，
随着1980年代苏联出现经济和货币危机，苏方要求朝鲜以市场价格向他们购买
粮食，加重了平壤政府的农业产业和粮食压力。1991年苏联解体，
朝鲜也因而失去了支援，经济因而崩溃，缺乏能力购入粮食和肥料……
1994年至1996年间，粮食产量锐减六成。

\casebar{Rank 2 --- More Translation-Useful Evidence}

\textbf{Similarity:} 0.859959

\vspace{0.5mm}
朝鲜地区产业以农业为主，耕地主要集中在西部沿海的平原和丘陵地区。
平安北道、平安南道、黄海北道和黄海南道（即``西海岸四省''）的平原地区
是朝鲜的``粮仓''。这里降水充沛，且地形平坦、灌溉系统相对完善，
主要用于高产的水稻和小麦集约化种植。咸镜道和江原道的沿海狭长平原
虽然面积较小，但也构成了东部重要的耕地带。

\medskip

\casefield{Analysis}{
The two candidates receive almost identical similarity scores, and the
top-ranked document even explicitly begins with ``朝鲜的农业''. However,
it mainly explains the historical causes of North Korea's food crisis rather
than its agricultural geography. The second document has slightly lower
similarity but directly describes the west-coast plains, the four provinces,
irrigation conditions, and the eastern coastal farming regions---the
information needed by the source sentence. Thus, stronger lexical matching
does not necessarily imply greater translation utility.
}

\end{casebox}

\begin{casebox}{Case 15: A Narrower Concept Can Dominate Similarity}

\casefield{Source}{
In physics and thermodynamics, an equation of state is a thermodynamic
equation relating state variables which describe the state of matter under
a given set of physical conditions, such as pressure, volume, temperature,
or internal energy. Equations of state are useful in describing the
properties of fluids, mixtures of fluids, solids, and the interior of stars.
}

\casefield{Query}{
``状态方程''
}

\casebar{Rank 1 --- Higher Semantic Similarity}

\textbf{Similarity:} 0.874748

\vspace{0.5mm}
一定量处于平衡态的气体，其状态与压强、$V$和$T$有关，表达这几个量之间
关系的方程称为\textbf{气体的状态方程}，不同的气体有不同的状态方程。
但真实气体的状态方程通常十分复杂，而理想气体的状态方程具有非常简单的形式。
在普通状况下，大多数实际气体的物理行为近似于理想气体……

\casebar{Rank 2 --- More Translation-Useful Evidence}

\textbf{Similarity:} 0.871450

\vspace{0.5mm}
【状态方程】物态方程最显著的作用是它可以通过已知条件来预测气体和液体的
状态，是一个标准的热力学方程。具有这样目的的、较为简单的物态方程是
理想气体状态方程……目前，没有一个单独的方程可以精确地预测所有物质在
所有情况下的状态；然而，其能很大程度上帮助描述流体、流体混合物、
固体和恒星内部特性。

\medskip

\casefield{Analysis}{
The first candidate repeatedly matches the phrase ``状态方程'', giving it
the higher similarity score, but its content is specifically about
\emph{gas} equations of state. The source instead defines the general
thermodynamic concept across fluids, mixtures, solids, and stellar interiors.
The second candidate repeats the query less strongly but discusses precisely
this broader concept and even covers the same application domains as the
source. This again shows that semantic similarity can favor a lexically
salient but overly narrow document over evidence with greater downstream
translation utility.
}

\end{casebox}

\subsection{When BLEU Disagrees with Translation Quality}
\label{app:bleu-cases}

The following cases illustrate why the corpus-level BLEU results in the main
experiments should be interpreted together with semantic and error-based
metrics. A translation can closely reproduce reference surface forms while
mistranslating the central terminology.

\begin{casebox}{Case 10: Correct Terminology with Paraphrased Framing}

\casefield{Source}{
In go and ``shōgi'', a jōseki or jouseki is the studied sequences of moves
for which the result is considered ``balanced'' for both black and white
sides.
}

\casefield{Reference}{
在围棋和将棋中，定式是经过研究的招法序列，其结果被认为对黑白双方都是
``平衡''的。
}

\casebar{CSC}

在围棋和将棋中，角先或角先是指那些经过研究的棋局序列，其结果被认为对
黑白双方都是``平衡''的。

\casescore{
BLEU 68.63 \quad COMET 0.774 \quad GRF 0.682 \quad MQM $-10.0$
}

\casebar{\textsc{SRAT}}

在围棋和将棋中，定式（jōseki 或 jouseki）是指经过研究的走法序列，
其中双方（黑方和白方）的结果被认为是均衡的。

\casescore{
BLEU 42.94 \quad COMET 0.789 \quad GRF 0.970 \quad MQM $-1.0$
}

\medskip

\casefield{Analysis}{
CSC closely reproduces the reference's surrounding wording and therefore
receives much higher BLEU, but ``角先'' is not the correct Go term.
\textsc{SRAT} correctly uses ``定式'' while paraphrasing the surrounding
sentence, reversing the ranking under GRF and MQM.
}

\end{casebox}

\begin{casebox}{Case 11: Named-Entity Accuracy versus Surface Overlap}

\casefield{Source}{
The Epistle of Jude, often shortened to Jude, is the penultimate book of
the New Testament and the Bible as a whole and is traditionally attributed
to Jude, the servant of Jesus and the brother of James the Just.
}

\casefield{Reference}{
犹大书，通常简称为犹大，是新约和整本圣经的倒数第二卷书，传统上认为是
耶稣的仆人、雅各的兄弟犹大所写。
}

\casebar{CSC}

犹大书，通常简称为犹大，是新约圣经和整个圣经的倒数第二本书，
传统上归于耶稣的仆人、雅各布斯的兄弟犹大。

\casescore{
BLEU 66.71 \quad COMET 0.882 \quad GRF 0.880 \quad MQM $-9.0$
}

\casefield{Retrieved Evidence}{
\textsc{SRAT} queries ``犹大书'' and retrieves an article identifying the
author as ``耶稣基督的仆人，雅各的弟兄犹大''.
}

\casebar{\textsc{SRAT}}

《犹大书》（常简称为犹大）是新约圣经及整个《圣经》中的倒数第二卷书，
传统上认为该书由耶稣的仆人、也是雅各（义人雅各）的兄弟犹大所写。

\casescore{
BLEU 47.67 \quad COMET 0.894 \quad GRF 0.973 \quad MQM $-1.1$
}

\medskip

\casefield{Analysis}{
CSC gains high lexical overlap but invents ``雅各布斯'' for James the
Just. \textsc{SRAT} grounds the entity using retrieved evidence and obtains
substantially stronger semantic and MQM scores despite lower BLEU.
}

\end{casebox}

\subsection{How Black-Box Translators Use Retrieved Evidence}
\label{app:blackbox-cases}

Finally, we compare Claude and GLM 5 under the same retrieved context.
These cases help explain why the gains of \textsc{SRAT}-GLM are smaller than
those of \textsc{SRAT}-Claude in Appendix~\ref{app:glm}.

\begin{casebox}{Case 12: Rejecting Off-Topic Evidence}

\casefield{Source}{
Null modem is a communication method to directly connect two DTEs using an
RS-232 serial cable.
}

\casefield{Reference}{
零调制解调器是一种使用RS-232串行电缆直接连接两个DTE的通信方法。
}

\casefield{Retrieved Evidence}{
Both systems receive an irrelevant passage about the Unix null device
\texttt{/dev/null}, described as ``空设备'', rather than the networking
concept \emph{null modem}.
}

\casebar{GLM 5}

空调制解调器是一种使用RS-232串行电缆直接连接两个DTE的通信方式。
该名称源于历史上使用RS-232电缆连接两台电传打字机设备或两台调制解调器
以进行相互通信。

\casescore{
BLEU 68.01 \quad COMET 0.836 \quad GRF 0.920 \quad MQM $-9.0$
}

\casebar{Claude Sonnet 4.6}

零调制解调器是一种使用RS-232串行电缆直接连接两个DTE的通信方法。
这个名称源于历史上使用RS-232电缆连接两个电传打字机设备或两个调制解调器
以便相互通信。

\casescore{
BLEU 67.80 \quad COMET 0.865 \quad GRF 0.970 \quad MQM $-1.0$
}

\medskip

\casefield{Analysis}{
GLM allows the irrelevant passage to overwrite the correct term,
changing ``零调制解调器'' to ``空调制解调器''. Claude largely ignores
the off-topic evidence and preserves the canonical terminology. The
difference is therefore not retrieval itself, but the translator's ability
to reject irrelevant context.
}

\end{casebox}

\begin{casebox}{Case 13: Selective versus Indiscriminate Evidence Use}

\casefield{Source}{
Birmingham City Football Club is a professional football club in Birmingham,
England. Formed in 1875 as Small Heath Alliance, it was renamed Small Heath
in 1888, Birmingham in 1905, and Birmingham City in 1943.
}

\casefield{Reference}{
伯明翰城足球俱乐部是英格兰伯明翰的一家职业足球俱乐部。该俱乐部成立于
1875年，当时名为小希斯联队，1888年更名为小希斯，1905年更名为伯明翰，
1943年更名为伯明翰城。
}

\casefield{Retrieved Evidence}{
Both systems retrieve an article on ``伯明翰足球会'', which correctly
provides the founding name ``小希斯联队'' but also uses a different
regional writing style.
}

\casebar{GLM 5}

伯明翰城足球会是位于英格兰伯明翰的职业足球会。该队成立于1875年，
初名「小希斯联队」，1888年更名为「小希斯」，1905年更名为「伯明翰」，
1943年正式定名为「伯明翰城」。

\casescore{
BLEU 47.38 \quad COMET 0.908 \quad GRF 0.970 \quad MQM $-3.0$
}

\casebar{Claude Sonnet 4.6}

伯明翰城足球俱乐部是位于英格兰伯明翰的一家职业足球俱乐部。
该俱乐部成立于1875年，当时名为小希斯联队，1888年更名为小希斯，
1905年更名为伯明翰，1943年更名为伯明翰城。

\casescore{
BLEU 74.36 \quad COMET 0.943 \quad GRF 0.990 \quad MQM $0.0$
}

\medskip

\casefield{Analysis}{
Both translators extract the useful founding name. GLM, however, also
copies the retrieved document's regional register, wording, and punctuation,
whereas Claude incorporates the needed fact while largely preserving its own
translation. This illustrates translator-dependent ability to use retrieved
evidence selectively.
}

\end{casebox}

\end{CJK*}

\section{LLM-as-Judge Evaluation Templates}
\label{app:judge-prompts}

For reproducibility, we provide the complete LLM-as-Judge templates used in
our experiments. I3 follows the original IdiomKB evaluation
\citep{li-etal-2024-idiomkb}, while I5 extends it to a finer-grained
five-level scale. MQM follows the ThinMQM template
\citep{zhan2026large}. GRF and GRB follow the reference-free and
reference-based evaluation templates of RAGtrans
\citep{wang-etal-2025-retrieval}. For GRF and GRB, we divide the raw
0--100 score by 100 when reporting results.


\begin{promptbox}{Idiom-Judge-3 (I3)}

Evaluate the idiom translation in the given Chinese translation of an English
sentence. Focus on the idiom's figurative meaning.

/* Evaluation Criteria */

1 point: Ignores, mistranslates, or only translates the literal meaning of the
idiom.

2 points: Conveys basic figurative meaning but may lack refinement or have
minor imperfections.

3 points: Exceptional translation, accurately conveying figurative meaning,
context, and cultural nuances.

/* Test Data */

Evaluate the following translation:

English sentence: \{src\}

Idiom in the English sentence: \{idiom\}

Chinese translation: \{mt\}

Evaluation (score only):

\end{promptbox}


\begin{promptbox}{Idiom-Judge-5 (I5)}

Evaluate the idiom translation in the given \{tgt\_lang\} translation of a
\{src\_lang\} sentence. Focus on the idiom's figurative meaning.

/* Evaluation Criteria */

1 point: Ignores or mistranslates the idiom, or renders only its literal
meaning; the figurative sense is lost.

2 points: Attempts the figurative meaning but gets it largely wrong or
distorted; significant inaccuracy.

3 points: Conveys the basic figurative meaning but lacks refinement, or has
minor inaccuracies or awkward phrasing.

4 points: Accurately conveys the figurative meaning with good fluency; only
slight imperfections in nuance or naturalness.

5 points: Exceptional translation, accurately conveying figurative meaning,
context, and cultural nuances with natural, idiomatic phrasing.

/* Test Data */

Evaluate the following translation:

\{src\_lang\} sentence: \{src\}

Idiom in the \{src\_lang\} sentence: \{idiom\}

\{tgt\_lang\} translation: \{mt\}

Evaluation (score only):

\end{promptbox}

\begin{promptbox}{GRF: Reference-Free Judge}

Score the following translation from English to [Chinese/German] on a
continuous scale from 0 to 100, where score of zero means "no meaning
preserved" and score of one hundred means "perfect meaning and grammar".

English source: [src]

[Chinese/German] translation: [hyp]

Score:

\end{promptbox}

\begin{promptbox}{GRB: Reference-Based Judge}

Score the following translation from English to [Chinese/German] with respect
to the human reference on a continuous scale from 0 to 100, where score of
zero means "no meaning preserved" and score of one hundred means "perfect
meaning and grammar".

English source: [src]

[Chinese/German] human reference: [ref]

[Chinese/German] translation: [hyp]

Score:

\end{promptbox}


\begin{promptbox}{MQM (ThinMQM)}

INPUT: \{Source Language\}

source: \{Source Text\}

\{Target Language\}

translation: \{Translation Text\}

Based on the source segment and machine translation surrounded with triple
backticks, identify error types in the translation, classify them and give a
score.

The categories of errors are: accuracy (addition, mistranslation, omission,
untranslated text), fluency (character encoding, grammar, inconsistency,
punctuation, register, spelling), style (awkward), terminology (inappropriate
for context, inconsistent use), non-translation, other, or no-error.

Each error is classified as one of three categories: critical, major, and
minor. Critical errors inhibit comprehension of the text. Major errors disrupt
the flow, but what the text is trying to say is still understandable. Minor
errors are technically errors, but do not disrupt the flow or hinder
comprehension.

For each error span, please deduct corresponding points based on following
reference points:

-25 = "Critical"

-5 = "Major"

-1 = "Minor"

-0.1 = "Minor/Fluency/Punctuation:0.1"

0 = "No-error"

OUTPUT:

Okay, let's tackle this translation quality assessment task. First, I need to
analyze the translation and classify the errors.

Critical: \{Critical Error ESA\}

Major: \{Major Error ESA\}

Minor: \{Minor Error ESA\}

Finally, I can calculate the final score:
\{Score Calculation Process\} </think>

Score: \{Final Score\}

\end{promptbox}

\end{document}